\documentclass{article}
\usepackage[a4paper, margin=2.78cm]{geometry}

\usepackage{graphicx}
\usepackage{xcolor}
\usepackage{authblk}
\usepackage{caption}
\usepackage{subcaption}
\usepackage{algorithm}
\usepackage{amsmath}
\usepackage{booktabs}
\usepackage{url}
\usepackage{hyperref}
\usepackage{amsthm}
\usepackage{amssymb}

\theoremstyle{definition}
\newtheorem{definition}{Definition}[section]
\theoremstyle{example}
\newtheorem{example}{Example}[section]

\newcommand{\fref}[1]{(\protect\subref{#1})}

\newcommand{\yp}{\widehat{y}}
\newcommand{\xp}{\widehat{y}}
\newcommand{\argmax}{\operatornamewithlimits{argmax}}
\newcommand{\R}{\mathbb{R}}
\newcommand{\Exp}{\mathbb{E}}
\renewcommand{\emph}[1]{\textit{#1}}

\begin{document}

\title{Supervised Learning from Data Streams:\\ An Overview and Update}

\author[1]{Jesse Read}
\author[2]{Indrė Žliobaitė}

\affil[1]{\small LIX, École Polytechnique, IP-Paris, France}
\affil[2]{\small University of Helsinki, Finland}

%\author{Jesse Read\thanks{LIX, Ecole Polytechnique} and \thanks{University of Helsinki}}
%\email{jesse.read@polytechinque.edu}
%\orcid{0000-0002-1013-6724}
%\affiliation{
%  \institution{LIX, Ecole Polytechnique}
%  \country{France}
%}

%\email{indre.zliobaite@helsinki.fi}
%\orcid{0000-0003-2427-5407}
%\affiliation{
%  \institution{University of Helsinki}
%  \country{Finland}
%}

\maketitle

\begin{abstract}
The literature on machine learning in the context of data streams is vast and growing. This indicates not only an ongoing interest, but also an ongoing need for a synthesis of new developments in this area. Here, we reformulate the definitions of supervised data-stream learning, alongside consideration of contemporary concept drift and temporal dependence. Equipped with this, we carry out a fresh discussion of what constitutes a supervised data-stream learning task; including continual and reinforcement learning; highlighting major assumptions and constraints. We carry out a fresh reconsideration of approaches and methods with regard to their suitability to modern settings. But more than a categorization of state-of-the-art streaming methods, we provide a reintroduction to what is supervised stream learning, and our emphasis here is a survey of settings. Our main goal is to pull theory and practice of supervised learning over data streams closer together. We conclude that practical stream learning does not mandate an online-learning regime. In the modern context, learning regimes should be selected and developed according to data-arrival modes, resource constraints, and maximum robustness and trustworthiness. We finish with a set of recommendations to this effect.
\end{abstract}
\section{Introduction}
\label{sec:intro}

A large area of the machine learning research literature deals with supervised learning in the context of data streams \cite{MOAbook,Gama_book,Aggarwal_book}, often focused on classification; occasionally also regression. This scenario involves
learning and prediction in an environment where instances are arriving continuously;
rather than being supplied as a fixed dataset prior to model deployment as in the classical machine learning.

Indeed, published in the last five years are around 20k papers that mention classification and data streams in the title\footnote{Google Scholar, 2025}. Many of these studies propose methods designed for or tested in sandbox settings that are sometimes only loosely connected to everyday real-world challenges. These papers commonly cite the ubiquity of streams in the real world, yet typically only test their methods on synthetic or static data sets which are at best small snapshots of past streams or are artificially turned into streams by manual sorting\footnote{For example, the Forest Covertype or Poker Hand datasets \cite{MOAbook}.}. Many of the popular benchmark sets are decades old and have relatively small dimensions. 

While not exclusive to data streams, concerning is that research papers often single out and apply assumptions and constraints without challenging their realism, or utility of the task. 
For instance, online and incremental learning regimes are often posited as mandatory due to a presumed impossibility or poor practice of storing instances, while excessive computational resources, to fit increasingly complex model structures or larger ensembles, are assumed to be available. Accurate drift detection is often prioritized without a study or consideration of expected drift patterns or change points. Immediate availability of ground truth labels is frequently assumed in research, while in reality that is rarely a possibility. We have used these assumptions in research ourselves. Although they often make scientifically interesting pilot studies feasible and fast, it is overdue to consider how best to go beyond pilot studies.

Here, we review and revise the assumptions and constraints behind supervised learning over data streams focusing on practical considerations in contrast to the academic preferences. We draw connections between operational choices and theoretical foundations. We also integrate related settings used across different research communities. The goal of our review is to provide a synthesis for setting in context and interpreting existing methods, and developing new modern solutions.

Without a doubt, we appreciate the importance of data-stream research contributions over many years. A great richness of empirical, methodological, and theoretical developments in the data stream learning research community have provided key scientific advancements in this area. 

As for any maturing discipline, the time comes for a synthesis, reflection on the academic research developments and potential ways forward. 
We provide revised definitions associated with data-stream learning tasks, and survey where we can find such tasks in practice, as well as which kinds of machine learning methodology are best applicable to such settings. 

Our main perspective is to treat the data generation and the instance delivery processes separately, and analyze how learning from data streams settings depend on both.

\section{Data Streams: Main Terminology and Definitions} 
\label{sec:fundamentals}

We begin by revising and main terminology associated with data streams.

\subsection{Preliminaries: Data and Concepts}

We denote random variable $X$ as associated with distribution $P(X)$,
\[
	x \sim P(X),
\]
which generates a data point $x$. This $P(X)$ is the generating distribution or \textbf{concept}. It corresponds to some real world process. 

\begin{definition}[Concept]
A concept is a category of meaning that a model can represent and reason about from examples. $P$ is the mathematical abstraction of the generating process. 
\end{definition}
From this process, we can draw or generate a number of observations 
\(
	x_1,x_2,\ldots,x_t,\ldots
\)

	where each \textbf{instance} $x_t$ is, thus, one observation of a real life process
    where $t$ denotes the place of this observation in the sequence of observations of interest for a given analytical task. For example, we flip a coin 10 times (thus $x_t$ takes a value either $\textsf{heads}$ or $\textsf{tails}$) producing data $x_1,x_2,\ldots,x_t,\ldots,x_{10}$. 

In this example, each $x_t$ can be represented by a binary number.
In other contexts, $x$ may represent a vector of attributes, a set, matrix, graph embedding, or other structured object.

This paper focuses on \emph{supervised} learning from data streams. We have \textbf{training examples} consisting of two parts $(x_t,y_t)$, where $y$ is a possibly multidimensional training label associated with a possibly multidimensional instance $x$ at time step $t$. 
The shared index $t$ does not necessarily imply that $x_t$ and $y_t$ were generated at the same time or place, but that they are a matching pair for learning purposes. For example, where $x_t$ may be the average temperature yesterday, and $y_t$ the average temperature today. Or, $x_t$ and $y_t$ may be the number of bicycles passing through two different areas of a city.

In the supervised learning setting we can denote a concept as 
\begin{equation}
	\label{eq:Pxy}
        (x,y) \sim P(X,Y).
\end{equation}

 Since $P(x,y) = P(y|x)P(x)$, then $P(y|x)$ is a part of the concept.
 
Each $(x_t,y_t)$-pair comes from the same distribution $P(X_t,Y_t)$,

representing a
real-world process. 

In supervised learning we seek a model that will map $x_t \mapsto \yp_t$, i.e., prediction $\yp_t$, such that a given loss metric $L(\yp_t,y_t)$ is low (equivalently: the predictive accuracy is high). 

\begin{figure}[!ht]
	\centering
	\begin{subfigure}[b]{0.45\textwidth}
		\centering
		\includegraphics[scale=0.8]{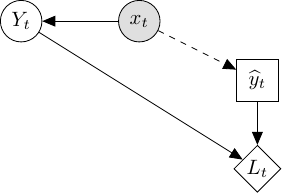} 
		\caption{Prediction context at time $t$}
		\label{fig:MP_IID.a}
	\end{subfigure}
	\begin{subfigure}[b]{0.45\textwidth}
		\centering
		\includegraphics[scale=0.8]{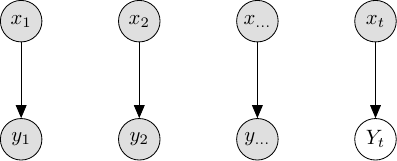}

		\caption{Unrolled version of \fref{fig:MP_IID.a}.}
		\label{fig:MP_IID.b}
	\end{subfigure}
	\caption{\label{fig:MP_IID}The supervised learning task over data generated from the same underlying distribution (concept): \fref{fig:MP_IID.a} as an influence diagram (circle nodes are random variables, square node a decision, diamond node a deterministic loss function) \cite{Barber}. This panel exemplifies concept $P(Y_t, x_t) = P(Y_t \mid x_t)P(x_t)$. 
 The link from $x_t$ to $y_t$ corresponds to $P(Y_t | x_t)$. In \fref{fig:MP_IID.b} this same concept is unrolled from $t=1$ in the fully supervised data-stream context (gray shading indicates observed data; transparent background denotes unobserved variables). 
 } 
\end{figure}

 Fig.~\ref{fig:MP_IID} depicts an abstraction of $P$ as a probabilistic graphical model. Panel~\fref{fig:MP_IID.a} shows a diagram for information related to prediction at time $t$. The value of $y_t$ is unknown at time $t$. Assuming immediate arrival of the ground truth label $y_{t-1}$ is known at time $t$ and thus loss $L$ at time $t$ is also known. Panel \fref{fig:MP_IID.b} shows the availability of instances over time. Depending on what is the current time point, the same variable can be in a known or an unknown state. Thus the figure emphasizes the statistical association (not necessarily causality) between an instance and its label. The learning and the prediction tasks are connected so that as time progresses, and the connection between the learning and prediction tasks.

\subsection{Data streams}
\label{sec:data_streams}

The key element which makes data stream learning a distinct area of study is the way data is accessed. We refer to this process as \emph{delivery} of data. A delivery process is responsible for transforming data into instances for learning, assigning time indices, and delivering these instances to other processes inside a software or hardware system. The delivery process is different from pre-processing the traditional data analysis, since it happens at the same time as the learning rather than prior to learning. In addition, the traditional view on pre-processing focuses on design choices, while here we emphasize the timing of those actions. The delivery process in data streams, thus, relates to the mode of data access rather than the data representation. The temporal nature of access defines data streams.

\begin{definition}[Data stream]
	A data stream is a mode of access  to data via
    a potentially-infinite sequence of observations; providing access to data in the form of instance $x_t$ uniquely at time step $t$ which denotes the time of access or delivery to a learning algorithm.

\end{definition}

\begin{figure}[!ht]
	\centering
	\includegraphics[width=0.50\columnwidth]{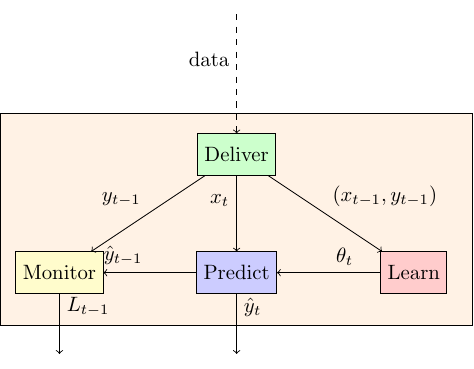}
	\caption{\label{fig:mainflow} Modules of an analytical system for learning and predicting from data streams; $t$ is current time. Arrows indicate actions happening at the current time step.}
\end{figure}

Fig.~\ref{fig:mainflow} illustrates the place of the Delivery process between the data source and other modules (such as Learning, Prediction, and Monitoring) of an analytical system. The figure illustrates the system at the current time step $t$. While Fig.~\ref{fig:MP_IID} which shows the mathematical abstraction, Fig.~\ref{fig:mainflow} relates to implementation and action. The following actions happen. Data arrives (dashed arrow) to the system. The Delivery module can pre-process the data if needed.
The Delivery process provides a time-ordered view of that data in the form of indexed instances to the \textsf{Learn}ing, \textsf{Prediction} and \textsf{Monitor}ing modules.
Downstream processes can never control an upstream mechanism. For example, a Learning module cannot send messages upstream to the Delivery mechanism.
The practitioner has complete control of anything implemented in the system. Thus, the Delivery is an issue of data management within a software or hardware implementation, as much as the Learning and Prediction is.

The modular learning system depicted in Figure Fig.~\ref{fig:mainflow}

produces multiple streams: a stream of instances for learning $x_1,x_2,\ldots,x_t,\ldots$, a stream of predictions $\yp_1,\yp_2,\ldots,\yp_t,\ldots$, a stream of model parameters $\theta_1,\theta_2,\ldots,\theta_{t-1}$, and a stream of the ground true labels $y_1,y_2,\ldots,y_{t-1}\ldots$ and a stream of loss measures $L_1,L_2,\ldots,L_{t-1},\ldots$. The stream of losses may be relevant for monitoring the system performance and, if enabled, drift detection. Drift detection is an unsupervised learning task, which is out of the scope of this review. More details on change detection can be found in \cite{Basseville_Nikoforov,IndresSurvey,Haug22} and in Section~\ref{sec:drift_detection_matter}. Note that we can generalize to $t-l$ where $l \geq 1$ is the lag for arrival of the true labels. In the following Example \ref{ex:temp}, $l=3$.

\begin{example}[Time series stream]
    \label{ex:temp}
	Temperature readings are obtained from a sensor at regular intervals. At each time point the \textsf{Delivery} process forms an instance consisting of the current and the previous observation  and delivers it to the Prediction process. The prediction target is to predict the temperature three time intervals ahead.
    Suppose we observe a sequence of readings
    ${\small \ldots, 13.5, 12.8, 14.4, 16.2, 19.9}$, the last observation is at time step $6$. The following instance can be formed for learning:
    $x_2 = (13.5, 12.8)$, $y_2 = 19.9$. 
    The following instance can be formed for prediction:
    $x_5 = (16.2, 19.9)$, $\yp_5 \in \R$. 

     Suppose Prediction $\yp_5 = 19.1$. If we use a quadratic loss function $L_{t} = (\yp_{t}-y_{t})^2$, then the loss that estimates the performance of the learning system at time $2$ is $L_2 = (19.1-19.9)^2 = 0.64$ .

	Note that loss can only be computed at the time (in this example, $t=5$) when the true temperature reading for the predicted horizon becomes available.
\end{example}

\begin{definition}[Time series]
\label{def:time_series}
	A time series is a sequence of observations $x_t$ indexed by time stamps $t$; where $x$ can be multivariate.
\end{definition}

The main difference between a time series and a data stream is that a time series is indexed by time stamps which may differ from the time of arrival of this data point to the learning process. In data streams index $t$ indicates the order in which $x_t$ is delivered downstream. Thus, a time series is a data structure, which can be analyzed sequentially or all the time-indexed observations can be analyzed at once, whereas a data stream is a mode of data arrival.

Data streams can be formed from time series data, but not every data stream is a time series and not every time series is a data stream.

The Delivery process in data streams may modify not only the order, but also the format of instances. Indeed, any data stream process module (Deliver, Prediction, Learning) can be seen as an operator on its input stream.

Below, in Fig.~\ref{fig:chess} we provide an example where data from time series involving temporal dependence via Delivery process that samples observation in random order is converted into a concept where no temporal dependence is observable.

Not only the Delivery process, which is at least partially in control of the user, can modify the data distribution. The data distribution may change over time outside the learning system. The phenomena is known in the data-stream machine-learning literature as concept drift.

\subsection{Concept drift}
\label{sec:what_is_concept_drift}

The topic of concept drift is tightly although not exclusively linked to data stream learning.
The understanding of what exactly constitutes concept drift and how best to deal with it is a subject of ongoing research \cite{Widmer96,IndresSurvey,webb2016characterizing,Lu19,Disabato22,vela2022temporal}. 

\begin{definition}[Concept drift]
	\label{def:drift}
	Concept drift is an unpredictable change to the concept (data distribution) over time, such that, for at least some $t$: 
\begin{equation}
	\label{eq:CD}
	P_{t-1} \neq P_{t}
\end{equation}
\end{definition}
(and, equivalently: $P(X_{t-1},Y_{t-1}) \neq P(X_t,Y_t)$). Thus, concept drift is when data points are \emph{not distributed identically} across a stream.

One can envision a sequence of data distributions $P_1,P_2,\ldots,P_t,\ldots$, such that 
\begin{equation}
	\label{eq:D}
	P_t \sim D(\mathcal{P}) 
\end{equation}
where $\mathcal{P}$ is a random variable of which the support is the set of \emph{all possible concepts}. However, in typical machine learning tasks no concrete expression for any of these concepts is available to the user, the task of detecting and reacting to concept drift is often part of Learning over data streams. 

\begin{example}[Predicting the demand for public transport]
	\label{eg}
	Let $(x_t,y_t) \sim P_t$ be a concept describing the hour of the day ($x$) and the number of public transport passengers using a particular bus line in a city during period of time marked by $t$ (an hour of day). At time $t+1$ a major accident happens that requires closure of a bridge for many months and the bus line has to make a long detour, which changes the passenger flows. From that point onwards observations start coming from a different concept such that $P_{t+1} \neq P_t$. The change would not have been predictable from the point of view of the learning module.
\end{example}

A concept $P$ can be decomposed following the joint rule of probability as 
\begin{equation}
	\label{eq:joint_rule}
	P(x,y) = P(x|y)P(y) = P(y|x)P(x)
\end{equation}
Any of these components may be subject to drift. If $x$ or $y$ are multidimensional, all the features may drift or only a subset of features may drift.  A change in $P(x)$ over time can happen so that $P(y|x)$ is not affected, this is known as virtual drift \cite{Tsymbal04}. Real drift, on the other hand (directly affecting the optimal decision surfaces of models) is a change in $P(y|x)$, which may or may not manifest as a change in $P(x)$ \cite{Tsymbal04,Zliobaite10detectable}. 

Concept drift can be categorized in terms of length (over time), its magnitude, and pattern \cite{IndresSurvey}. The magnitude refers to the distance between $P_t$ and $P_{t+1}$ (in concept space $\mathcal{P}$), and the configuration refers to how frequently $P_{t} \neq P_{t+1}$ (versus how often the two distributions are equal). 

 In these two dimensions, drift patterns can be described as gradual, abrupt, incremental, seasonal or anomalous \cite{IndresSurvey}. Since an infinite number of drift types can be conceived, drift categorization is largely subjective yet it can serve as an empirical guideline when designing learning algorithms that can adapt to drifts. 

The literature is limited insofar as demonstrating what kinds of drift occur in practice. But it is generally accepted in the literature that the onset of drift is unforeseeable, and the behaviour of drift is unpredictable in the long term \cite{lu2018learning}. Interestingly, most dictionary definitions of the word `drift' refer to  a \emph{slow} process. This would suggest that most streams are dominated by stationary or almost-stationary concepts, which is rarely emphasised in data-streams research. 
Certainly, the terms such as `long term', and `slow' are subjective as well. 

When the concept is moving towards a known or predictable outcome, that is not considered as concept drift, since those changes can be deterministically inscribed into the data distribution $P$. For example, the regular annual seasonality of temperatures outside is generally not considered as concept drift, but rather part of a concept with temporal dependence. We discuss this next in Section~\ref{sec:temporal_dependence}. 

While the term \textit{concept drift} is strongly associated with supervised learning from data streams, changing data distribution occurs in several other settings of computational data analysis. Related phenomena are known as \textit{hidden contexts} \cite{Widmer96,Harries98}, that refer to changes in the distribution of latent (unobserved) variables over time. In the statistical learning community, the scenario when the joint distribution of inputs and outputs differs between training and testing stages is known as dataset shift \cite{quinonero2022dataset}, including distribution shift \cite{Mathelin25}, including, covariate shift. Covariate shift only involves change in the input data distribution $P(x)$, since covariate refers to the independent variable $x$ in the supervised learning setting \cite{Shimodaira00}. The deep learning community refers to the context of out-of-domain generalization and dataset shift \cite{wald2021calibration,gardner2023benchmarking,chang2025model}. Dataset shift (and other terms covering changes in concept) do not necessarily involve learning over streaming data. A learner is trained on a dataset from one distribution and applied to a dataset expected to be distributed differently. In those circumstances the point of shift is usually known (it is delimited by the change in dataset), even the magnitude of the shift may not be known. In the more general setting of transfer learning, not only the data distribution but also the data representation and task formulation may change as well \cite{Pan09}.

\subsection{Temporal dependence}
\label{sec:temporal_dependence}

Variables $X_i$ and $X_j$ are independent when $P(X_i,X_j) = P(X_i)P(X_j)$ \cite{Barber}. Temporal dependence arises when this property does not hold between observations at different times, leading to the following definition.

\begin{definition}[Temporal dependence]
	\label{def:temporal}
	A time series $x_1,x_2,x_3,\ldots,x_t$ exhibits temporal dependence when at least two observations $x_i$ and $x_j$ ($i\neq j$), are statistically dependent such that
	\begin{equation}
		\label{eq:neq}
		P(x_i | x_j) \neq P(x_i)
	\end{equation}
	meaning that the distribution of $x_i$ depends on $x_j$. 
\end{definition}

\begin{figure}[!ht]
	\centering
	\includegraphics[width=0.75\columnwidth]{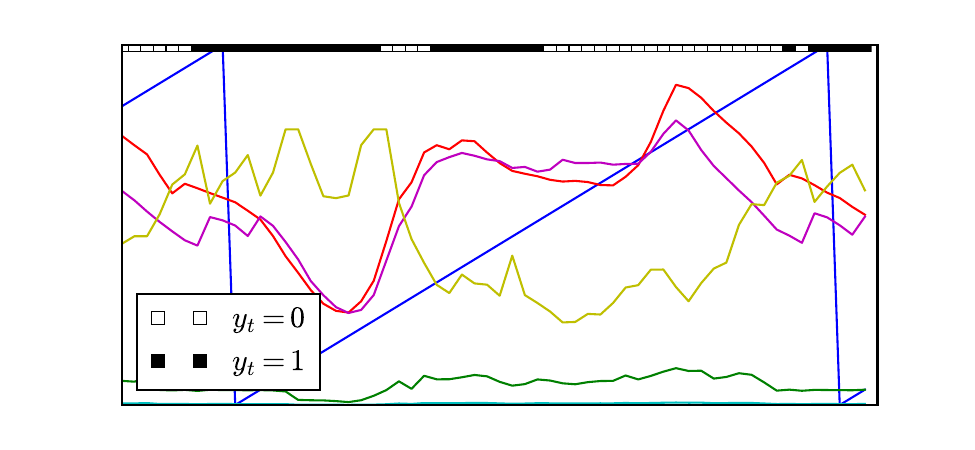}
	\caption{\label{fig:electricity}A snapshot of the Electricity dataset used as a benchmark in streams literature.}
\end{figure}

Fig.~\ref{fig:electricity} shows a snapshot of the Electricity dataset often used in streams literature \cite{Zliobaite13electricity}, exhibiting temporal dependence across $x_t$ and $y_t$. 

\begin{figure}
	\centering
	\begin{subfigure}[b]{0.45\textwidth}
		\centering
		  \includegraphics[width=1\columnwidth]{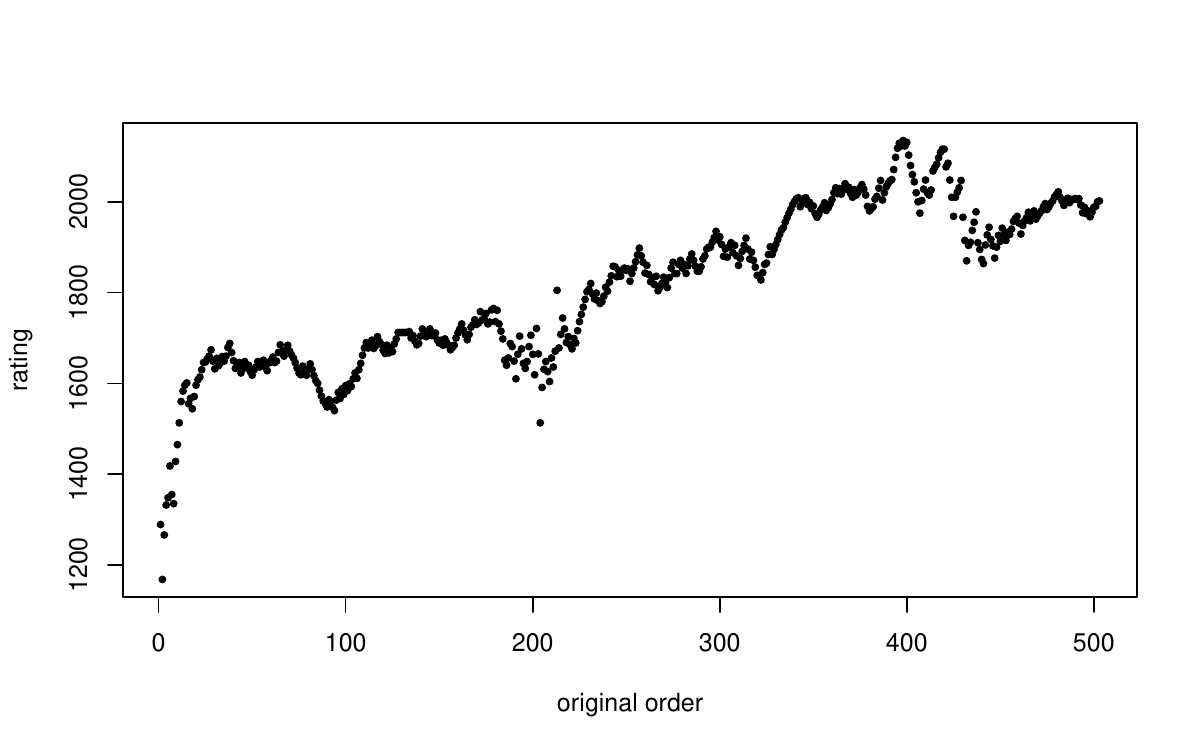}
		\caption{\label{fig:chess.a}Time series view}
	\end{subfigure}
	\begin{subfigure}[b]{0.45\textwidth}
		\centering
		  \includegraphics[width=1\columnwidth]{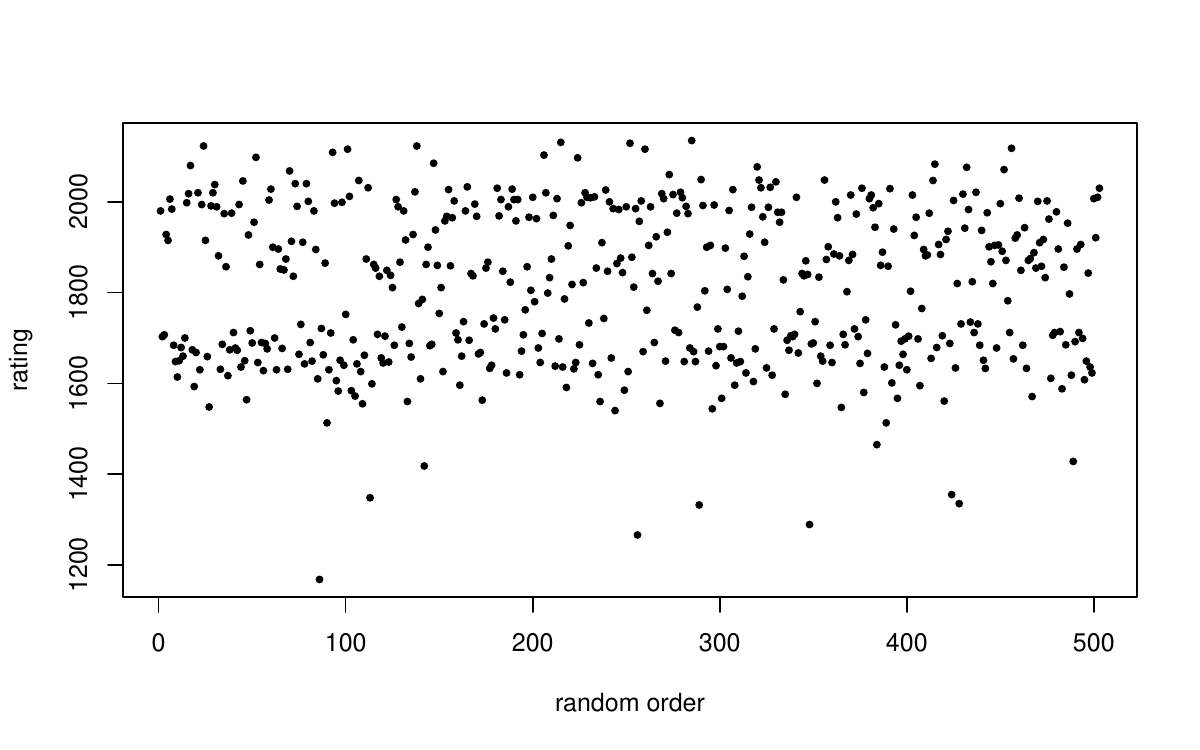}
		\caption{\label{fig:chess.b}Data-stream (random) mode of access}
	\end{subfigure}
	\caption{\label{fig:chess} Chess rating trends from \cite{Zliobaite11}: in \fref{fig:chess.a} original time series and \fref{fig:chess.b} a data stream with a random mode of access.}
\end{figure}

The Delivery process (recall, e.g., in Fig.~\ref{fig:mainflow}) can remove dependence, for example, by serving observations in a random order as exemplified in Fig.~\ref{fig:chess}. 

\begin{figure}[!ht]
	\centering
	\begin{subfigure}[b]{0.45\textwidth}
		\centering
		\includegraphics[scale=0.8]{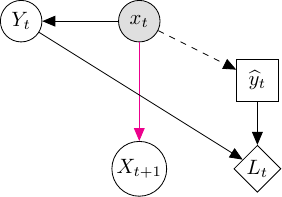} \quad\quad\quad
			\caption{Prediction context at time $t$.}
		\label{fig:MP_series.a}
	\end{subfigure}
	\begin{subfigure}[b]{0.45\textwidth}
		\centering
			\includegraphics[scale=0.8]{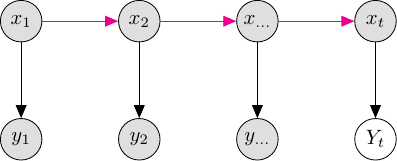}
		\caption{Unrolled version of \fref{fig:MP_series.a}.}
		\label{fig:MP_series.b}
	\end{subfigure}

	\caption{\label{fig:MP_series}
	A concept involving Markovian temporal dependence $P(X_t \mid x_{t-1}) \neq P(X_t)$ (shown in {\color{magenta}magenta}). If $Y_t$ is the target for prediction, this type of dependence has no effect on predictive strategy; unlike that shown in Fig.~\ref{fig:MP_series_imp}.}
\end{figure}

Fig.~\ref{fig:MP_series} shows a mathematical abstraction as a probabilistic graphical model for a data stream involving temporal dependence. It explicitly defines $P(X_t \mid x_{t-1})$ as part of the concept. 

\begin{figure}[!ht]
	\centering
	\includegraphics[scale=0.8]{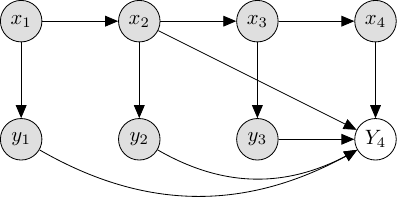}

	\caption{\label{fig:MP_series_imp}
	An example of temporal dependence where, unlike in Fig.~\ref{fig:MP_series.b}, it is necessary to make special algorithm considerations, namely take into account the observations $x_2,x_4$, $y_1,\ldots,y_3$, when predicting $Y_4$. A Delivery process can convert this concept into that of Fig.~\ref{fig:MP_series.b}, as described in Section~\ref{sec:data_streams}. 
	}
\end{figure}

Temporal dependence is not the same as concept drift. In case of temporal dependence it can be that $P_t(x_t) \neq P_t(x_t \mid x_{t-1})$; in the case of concept drift it can be that $P_t(x_t \mid x_{t-1}) \neq P_{t-1}(x_t \mid x_{t-1})$. In the latter, the concept $P_t$ itself (or some element thereof) changes. In the former, a concept is static, even when it is generating dynamic (temporally-dependent) data.

With regard to the assumption of identically and independently distributed (IID) data, concept drift violates the ``identical" assumption and temporal dependence violates the ``independence" assumption.

\subsection{Non-Stationarity vs Concept Drift}

Literature in statistics uses the term \textit{non-stationary} time series \cite{Hyndman_book,THEODORIDIS20159}, which is closely related to concept drift in data streams. 

The definition of a non-stationary time series data, is essentially the same as a concept-drifting stream: statistical properties of $P_t$, such as mean and variance, may change over time $t$. And a stationary time series, on the other hand, is one whose properties do not depend on the time at which the series is observed. Non-stationarity in time series can be equivalent to concept drift in data streams, when the stream provides a view over such a time series. 

\begin{figure}[!ht]
	\centering
	\begin{subfigure}[b]{0.4\textwidth}
		\centering
		  \includegraphics[width=1\columnwidth]{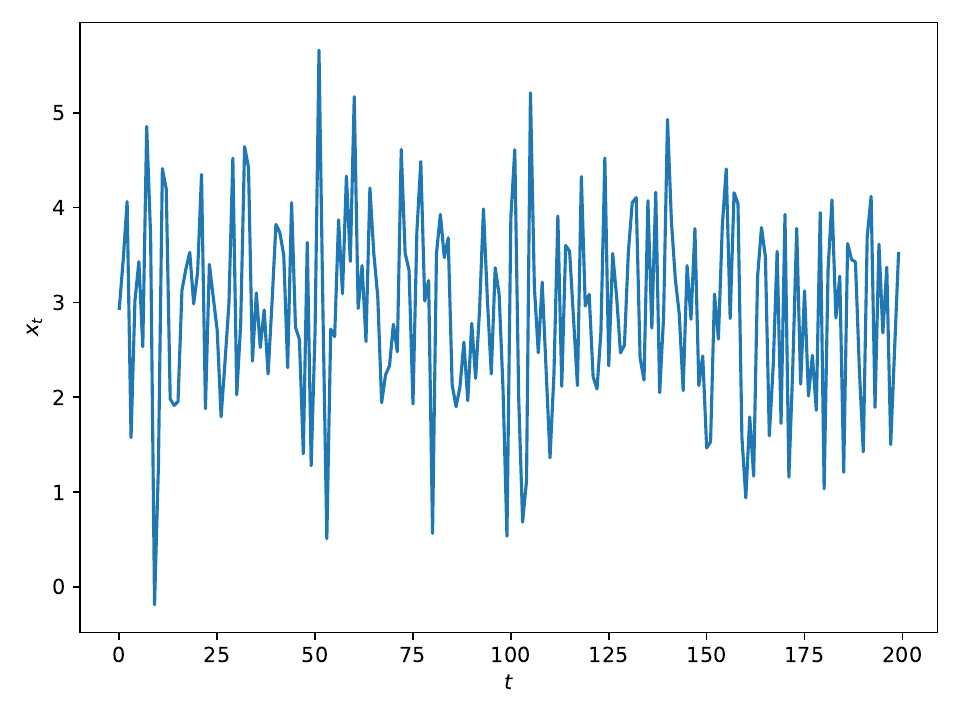}
		\caption{\label{fig:s.a}}
	\end{subfigure}
	\begin{subfigure}[b]{0.4\textwidth}
		\centering
		  \includegraphics[width=1\columnwidth]{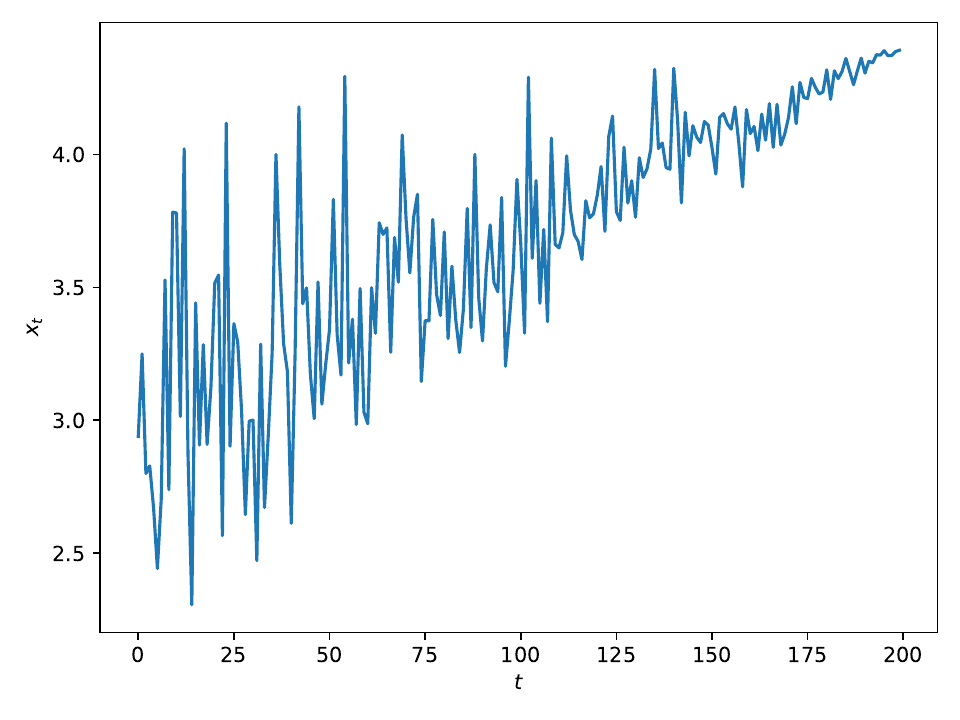}
		\caption{\label{fig:s.b}}
	\end{subfigure}
	\caption{\label{fig:stationary_vs_not}An example of \fref{fig:s.a} stationary vs \fref{fig:s.b} non-stationary time series; the difference can be seen for two reasons: changing mean (i.e., trend), and variance (i.e., properties of concept $P_t$, in relation to different windows over the data). }
\end{figure}

A synthetic example of non-stationary time series (vs stationary) is given in Fig.~\ref{fig:stationary_vs_not}.

However, there is a practical difference, linked to the fact that a stream provides a restricted view over `live' data (data given to a model under deployment, one instance at a time): non-stationarity implies a description of time-series data, whereas concept drift emphasises task relevance, and implications on model performance. Due to the restricted view of data offered by a stream, temporal trends implying non-stationarity can only be recognised retrospectively; they are unpredictable, and therefore considered as drift.

\section{Data Stream Learning: Task Settings}
\label{sec:task_settings}

In this section we first formalize the learning-task setting, starting with the generic case and then covering cases particularly relevant to streams, notably those involving temporal dependence. Later in Section~\ref{sec:methodology_and_strategies} we survey computational strategies for these tasks in terms of choice of model class, and learning algorithms to induce instantiations of such models from a data stream.

\subsection{Supervised learning from IID data}
\label{sec:iid}

In the classical supervised machine learning setting, we wish to use data set $\{(x_i,y_i)\}_{i=1}^{t-1}$, sourced from real-world concept $P$, to build a decision function (classifier or regressor) which provides predictions $\yp_t$ for any new instance $x_t$ that minimizes expected loss, 
\begin{equation}
	\label{eq:learn}
	\min_{\yp_t} \Exp_{Y \sim P(Y|x_t)}[L(Y,\yp_t)]
\end{equation}
for a user-defined loss metric $L$.

In the regression case, we typically denote the decision function $\yp = f_\theta(x)$ (let $\theta$ denote a particular instantiation of, or set of parameters defining, a model); and it provides a continuous-valued prediction. If the loss metric is squared-error, then $\yp$ should (to minimize Eq.~\eqref{eq:learn}) be the mean of distribution $P(Y|x)$, i.e.,  
\begin{equation}
	\label{eq:regression}
	\yp_t = f_\theta(x_t) = \Exp_{Y \sim P(Y|x_t)}[Y]
\end{equation}
In classification we typically denote function $\yp = h_\theta(x)$ which will return a categorical value. If the loss metric is error rate ($0/1$-loss or inverse accuracy), then this prediction should be the mode of the distribution, i.e.,

\begin{equation}
	\label{eq:provide}
	\yp_t = h_\theta(x_t) = \argmax_y P(y|x_t)
\end{equation}

Therein we see that the goal of learning is to capture some aspect of concept $P$, typically in a form of the conditional distribution  
\begin{equation}
	\label{eq:p_y}
	P(Y_t|x_t)
\end{equation}
illustrated in Fig.~\ref{fig:MP_IID} via probabilistic graphical models. Under the maximization (of Eq.~\eqref{eq:provide}) this is equivalent to estimating $P(x_t,Y_t)$, since $P(x_t,Y_t) \propto P(Y_t|x_t)$ and so $Y_t$.

We do not have access to any aspect of true concept $P$, and so must be able to infer and to update $\theta$ based on the available training data to potentially improve predictions. In data streams, the task is potentially complicated by concept drift, but the main goal remains the same: to build a model whose predictions minimize conditional expected loss.

So far, we have formalized the learning task in the standard assumption of independent and identically distributed (IID) data.

\subsection{Supervised learning from data streams that have temporal dependence}
\label{sec:temp_sec}

In reality, data from data streams is rarely or never IID \cite{AdversarialOnline,Electricity}. After surveying a wide range of potential stream-learning applications, a question arises of whether real-world supervised learning tasks over a naturally IID data streams exist, that is to say, where a data stream mode of delivery is realistic. We argue that most natural scenarios for supervised learning over data streams exhibit temporal dependence (as we have described in Section~\ref{sec:temporal_dependence}), violating the independence assumption of IID. The remainder of this section deals with such cases.

We emphasise the following: any data stream can be converted into an IID stream, implicitly or explicitly, by preprocessing and delivering instances in a form which `removes' this dependence. Obviously one can remove dependence simply by shuffling instances and delivering in a random order (consider: Fig.~\ref{fig:chess}) but here we may be losing information; therefore we do not recommend this approach under any circumstance of learning strategy. Rather, we can redefine what an  instance represents, removing inter-instance temporal dependence by capturing it within a single instance \cite{DuSwamy,Our2015Paper,ChainInTime,almeida2023time}. 

Fig.~\ref{fig:MP_series_imp} suggests that the delivery process could explicitly remove temporal dependence into a representation of an IID concept simply by redefining the input instance such that
\[
	x_4 \gets \underbrace{(x_2,x_4,y_1,y_2,y_3)}_{x_4}
\]
If this pattern of temporal dependence were stable (i.e., in the absence of concept drift) and limited to horizon 4 (as in the example), we can capture it generally with  
\[
	x_t \gets (\underbrace{x_{t-4},\ldots,x_t,y_{t-4},\ldots,y_{t-1}}_{\subset \{x_1,\ldots,x_t,y_1,\ldots,y_{t-1}\}})  
\]
i.e., a moving window of size 4; a subset of all \emph{observed} data. Whether or not all those observations are available, or if additional observations (from the future) are available, is an important question.

Table~\ref{tab:time_series} summarizes different supervised learning task formulations in terms of different configurations of observable input data and prediction targets.
Again we emphasise: all tasks listed in the table can be mapped to the standard IID learning tasks be redefining the inputs of the instance, either explicitly via the Delivery process or implicitly in the Learning process; in either case requiring some form of memory or buffer.

\begin{table}[!ht]
	\centering
 \caption{\label{tab:time_series}
	A summary of prediction tasks, showing the parts of a stream that may be used (when available) and may be required as a training label (target) to carry out optimal learning and decision making for the associated task in the context of temporal dependence. 
 }
		\begin{tabular}{|llll|}
			\hline
			\textbf{Type of modeling} & \textbf{Inputs} & \textbf{Target} & \textbf{Details} \\
			\hline
			 Classification and regression &  $x_t$ &  $y_t$ & Sec.~\ref{sec:iid} \\

			 Filtering & $x_1,\ldots,x_t,y_1,\ldots,y_{t-1}$    &  $y_t$   & Sec.~\ref{sec:filtering}, Fig.~\ref{fig:filtering}   \\
			 Forecasting ($l$-step ahead) & $x_1,\ldots,x_t,y_1,\ldots,y_t$    &  $y_{t+l}$    & Sec.~\ref{sec:forecasting}, Fig.~\ref{fig:forecasting}\\
			 Forecasting ($l$-step ahead) & $x_1,\ldots,x_t$    &  $x_{t+l}$    & Sec.~\ref{sec:forecasting}\\

			 Smoothing & $x_1,\ldots,x_t,\ldots,x_{t+l}$    &  $y_t$     & Sec.~\ref{sec:smoothing}, Fig.~\ref{fig:smoothing}\\
			 Autonomous agent (action $y_t$) & $x_1,\ldots,x_t$    &  $y_t$    & Sec.~\ref{sec:reinforcement}, Fig.~\ref{fig:reinforcement} \\
			 Bandit algorithm (action $y_t$) & $L_1,\ldots,L_t$    &  $y_t$    & Sec.~\ref{sec:bandits}, Fig.~\ref{fig:bandits} \\
			\hline
		\end{tabular}
\end{table}

\subsection{Filtering (nowcasting) with data streams}
\label{sec:filtering}
Filtering or nowcasting task refers to predicting the target value for the present time from observations in the past and present (thus implying that such observations are available).

The task of filtering is very well studied in signal processing from where it takes its name \cite{kalman1960new}, but the term is also well used in machine learning, where binary classification is sometimes called filtering \cite{guzella2009review}. The very first application of a multi-layer neural network was as a phone-line filter \cite{Widrow59}. In time series analysis this task is often called nowcasting, particularly in economics or meteorology \cite{khan2022nowcasting,coraddu2024floating}.

\begin{figure}[!ht]
	\centering

	\includegraphics[scale=0.5]{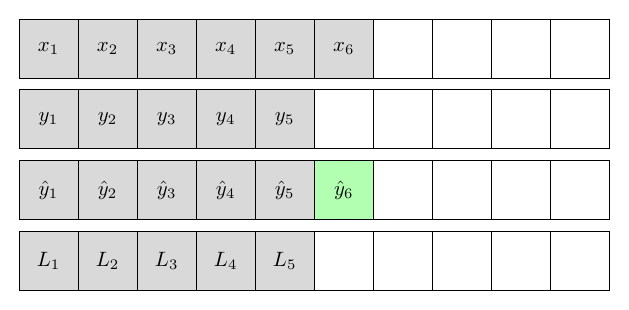}
	\caption{\label{fig:filtering} A filtering task in a stream; at current time $t=6$. A prediction must be provided, $\yp_6$. All elements shown in grey can be known at time $t=6$ to make this prediction. They may be unknown, for example, in cases of delayed labels or missing data.}
\end{figure}

Fig.~\ref{fig:filtering} gives an example of a data structure for the data-stream filtering task. The current time is assumed to be $t=6$. The objective is to output prediction $\yp_6$. Intuitively, $x_6$ passes through the filter to become $\yp_6$, an approximation of (\emph{unobserved}) $y_6$.

\begin{example}[Spam filtering]
    \label{ex:spam}
A spam classifier  \cite{sculley2007online,lughofer2017line,Dada19}
must decide 
if the current email $x_t$ is \textsf{spam} ($\yp_t=1$) or \textsf{not spam} ($\yp_t=0$), based on prior emails and their confirmed labels $(x_i,y_i)$, $i<t$. A confirmed label is when, for example, a user explicitly marks an email as spam ($y_i=1$), or replies to or labels an email thus confirming that it is not spam ($y_i=0$).

\end{example}

\subsection{Forecasting (predicting the future) with data streams}
\label{sec:forecasting}

Forecasting is a predictive modeling task where predictions about the future state of a target variable $y_{t+l}$ (or $x_{t+l}$) need to be made from observations available up to and including the present time. Here $t$ is the present time step, $l>0$ is the prediction horizon measured in time steps. Often, even not always, in this scenario the target value is a time-shifted value of the current observation, that is $y_t = x_{t+l}$.

Typically, this task only makes sense if temporal dependence (and thus relevant information for the future) is present. As in the filtering task, it can be that $x_t$ already carries sufficient temporal information to predict the target (i.e. a full history of $x$- and $y$-sequences may not be necessary or available), for example, in credit scoring, the information regarding the probability of repaying credit in the future may be predictable only from the current financial situation of a customer.

\begin{figure}[!ht]
	\centering

	\includegraphics[scale=0.5]{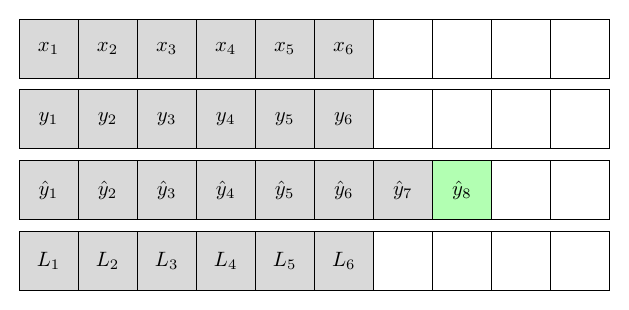}
	\caption{\label{fig:forecasting}A forecasting task in the stream setting at current time $t=6$. The task is to predict $\yp_6$. All elements shown in grey can be observed and can be used to make this prediction. The fact that $\yp_7$ is gray (observed) means it is possible to observe and use as evidence the model's predictions (prediction based on a prediction); this can make sense, since such predictions are estimating a trend, for example. 
	}
\end{figure}

An example of a data structure corresponding to the forecasting task is given in Fig.~\ref{fig:forecasting} for $l=2$ forecasting horizon. Variations are possible; for example, observations $x_t$ might not be available. 

\begin{example}[Electricity demand prediction]  At time $t$, we want to estimate electricity demand at $t+2$ (e.g., at time $t=6$, we produce estimate $\yp_8$; the forecast for future time $t=8$). In addition to past and current electricity usage, $y_1,\ldots,y_t$ we can also consider as input the outdoor temperature $x_1,\ldots,x_t$, over the same period. As part of the same, or a separate task, we might be interested in forecasting temperature observation $\xp_{t+2}$. 
\end{example}

Forecasting is one of the most suitable tasks of supervised learning from data streams. Applications include weather forecasting \cite{sanchez2019data}, traffic trajectory prediction \cite{ChainInTime}, energy systems \cite{fekri2021deep,melgar2021nearest}, epidemic forecasting \cite{desai2019real}, predictive maintenance or fault forecasting \cite{alzghoul2012data}; a number of other applications cited by \cite{lim2021time} in the context of deep learning.

We say that in practice many learning tasks over data streams are in the forecasting setting since only in this scenario true labels are easily available as the future becomes the present (even such labels arrive with a delay of $l$).

\subsection{Smoothing (nowcasting) with a delay}
\label{sec:smoothing}

The prediction for the $t$-th instance is not always required at the current time step $t$. It may be sufficient to provide the estimate at time step ${t+l}$. This task is often called \emph{smoothing} \cite{Barber}.
The option to predict with the largest allowable delay when affordable is expected to result in more accurate prediction performance \cite{Barber} because it allows to gather more data prior to committing to a prediction. In a sense it is the inverse of the forecasting problem. Inferring the past from the future can be an easier task than predicting the future, and this setting still pertains to many real-world tasks.

\begin{figure}[!ht]
	\centering

	\includegraphics[scale=0.5]{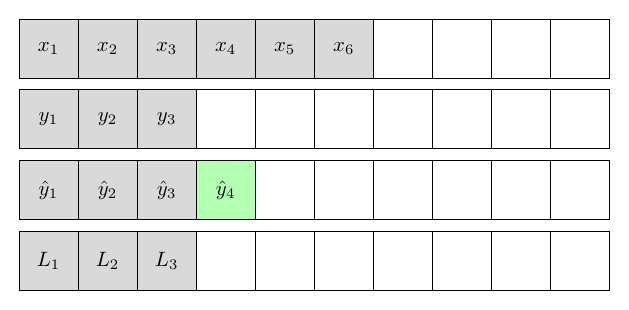}
	\caption{\label{fig:smoothing}A smoothing task over streaming data at $t=6$.
	Observed data is shown in shaded nodes, unobserved data in non-shaded nodes, and the prediction target node is squared. Since we are at time $t=6$, we have observed $x_6$; label $y_4$ was not yet observed; the task is to infer (predict) its value (note: lag $l=2$). }
\end{figure}

Fig.~\ref{fig:smoothing} illustrates the data stream setting and the data structure. 

\begin{example}[Time series stream]
    \label{ex:covid}
The number of positive COVID-19 test results $y_t$ in a country are recorded \cite{moran2016epidemic} across days $t=1,2,\ldots,t$. Records for some days are missing, for example, because those are public holidays and no official testing was carried out.  We would like to estimate the number of infections on those days retrospectively. We can use the existing $y$-measurements as well as additional data, denoted $x_1,x_2,\ldots,x_t$, such as observations on public transport use or self-reported results to make such an estimate.
\end{example}

Other examples include localisation of an object in a video stream which has been buffered $l$ frames ahead of current time $t$ under viewing; or a car in traffic with $l$-second delay. Advanced smoothing methods for these settings have been extensively researched \cite{nouvellet2021reduction}.

A special case of smoothing is known as early time series classification \cite{xing2012early}. 
Here a model is permitted to wait $l$ instances at a cost proportional to $l$. In this setting $l$ is part of the loss metric, $L(y,\yp,l)$, adding a penalty proportional to $|l|$. For example, a diagnosis for a patient needs to be made of their available tests up to time $t$. Waiting until $t+l$ for more test results can lead to a more accurate diagnosis, but also may lead to a deteriorating condition.

\subsection{Control systems and autonomous agents}
\label{sec:reinforcement}

Filtering, forecasting, and smoothing are all tasks where models are passive, in the sense that their estimates do not directly affect the future of the stream observations; they provide a new stream $\yp_1,\yp_2,\ldots,\yp_t,\ldots$, which has no effect on $x_{t+1}, x_{t+2}, \ldots$.

\begin{figure}[!ht]
	\centering
	\includegraphics[scale=0.8]{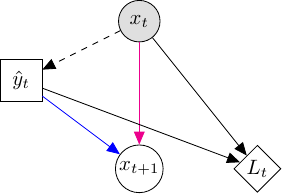}
	\caption{\label{fig:mdp}An influence diagram for a Markov decision process as typically considered in reinforcement learning. Here the label is at the same time a decision or action, which can affect future inputs (drawn in {\color{blue}blue}); as such, temporal dependence exists indirectly in the $x$-sequence, via model-outputs $\yp_t$.
	}
\end{figure}

In the context of control systems \cite{RLBook2018} the output $\yp_t$ \emph{is a decision or action that can affect the future stream}. In this setting, a model (known in reinforcement learning as a policy; and usually denoted $\pi$) influences
the future observations it receives; thereby having the ability and
the objective to minimize the accumulation of future losses. 
On account of this aspect, unlike the tasks of filtering, forecasting and smoothing, reinforcement learning cannot simply be reduced to supervised learning from IID data. Even when the instances are redefined to capture historical dependence, temporal dependence `leaks' through the model; seen clearly via the $\yp_t \mapsto x_{t+1}$ link in Figure~\ref{fig:mdp}. 

\begin{example}[Autonomous vehicles]
	An autonomous vehicle directs itself via actions $\yp_t$ (turn, brake, throttle, indicate, \ldots) through an urban environment. By taking certain actions, the vehicle avoids accidents and proceeds towards its next destination smoothly; not only incurring different observations $x_{t+l}$ than if it had acted otherwise, but and also incurring favourable (i.e., small) loss $L_{t+l}$ for various horizons of $l$. The task of a learning algorithm is to associate decisions and losses ($\yp_t$ vs $L_{t+l}$) across time (for potentially large $l$).
\end{example}

Other target applications include robotics \cite{ibarz2021train}, game-play, finance, control of energy systems \cite{AlbanTurbines} and finance (among many others).

\begin{figure}[!ht]
	\centering
	\includegraphics[scale=0.5]{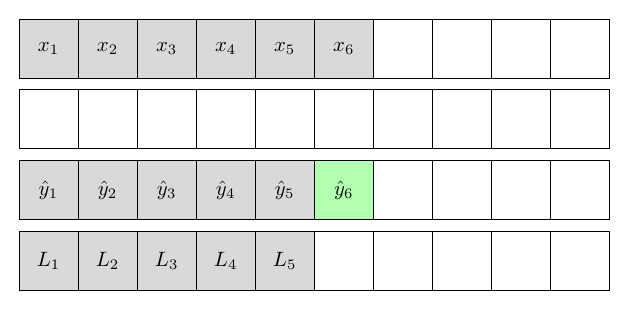}
	\caption{\label{fig:reinforcement}A Reinforcement Learning algorithm learns via direct interaction with an environment (concept $P(X_{t+1} \mid x_t, \yp_t$) and its decisions/actions ($\yp$) are reinforced by rewards (equivalently, losses $L$) that are linked to decisions after they are made. This is a streaming problem, but there is not a label $y_t$ available (except in the related task of imitation learning), and a decision $\yp_t$ may incur distant losses $L_{t+l}$ (for large $l$); hence the challenges involved. 
	} 
\end{figure}

Fig.~\ref{fig:reinforcement} illustrates the setting graphically, noting that $\yp_t$ may affect $x_{t+1}$ (and thus, indirectly, $x_{t+2}$, $L_{t+2}$, and so on ($t+l$). Notably absent are ground-truth actions $y_t$; hence the challenge of reinforcement learning. If such expert-provided actions are available, the task becomes imitation learning (specifically, behavioural cloning) \cite{zare2024survey}, which is a special case of filtering or smoothing (i.e., supervised learning), described above.

This learning setting frequently meets the typical assumptions held in the data-stream learning community regarding streams (always providing an output, always receiving instant feedback on a decision), yet so far reinforcement learning has not been thoroughly embraced by this same community upholding those assumptions.

Reinforcement learning uses the term \textit{environment} in a similar way to \textit{concept} in the context of the data stream literature; except the model is \emph{part of} this concept, since  $P(x_{t+1} \mid \yp_t, x_t)$. 

Concept drift in reinforcement learning corresponds to a change in $P$; but it is not possible for the model to influence this change.

For example, the decision of an autonomous car to drive faster affects its future observations and losses ($x_{>t}$ and $L_{>t}$, respectively) but not the physics of its environment ($P$).
However, if the environment is changed from the one in which it learned to drive, for example, under different weather conditions, then
a change in concept has occurred ($P$ has changed). In model-based reinforcement learning \cite{moerland2023model}, an additional stream-based goal is to infer $P$, i.e., to model the concept; this is analogous to explicitly learning a representation of a concept alongside a data stream-classifier.
Also in reinforcement learning, researchers are aware of the need to transfer knowledge to new environments (concepts) \cite{taylor2009transfer}, i.e., adaptation to concept drift.

\subsection{Bandit machines}
\label{sec:bandits}

\begin{figure}[!ht]
	\begin{subfigure}[b]{0.45\textwidth}
		\centering
	\includegraphics[scale=0.8]{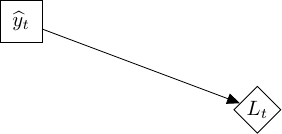}
		\caption{}
		\label{fig:bandits.a}
	\end{subfigure}
	\begin{subfigure}[b]{0.45\textwidth}
		\centering
		\includegraphics[scale=0.5]{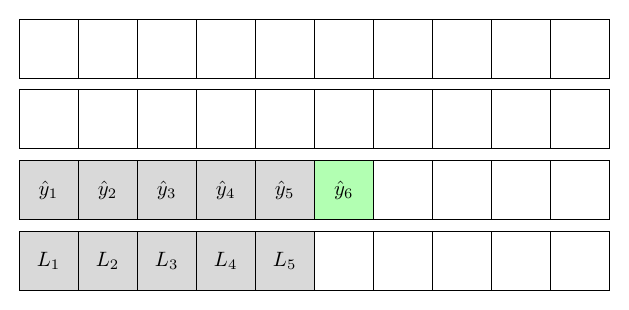}

		\caption{}
		\label{fig:bandits.b}
	\end{subfigure}
	\centering
	\caption{\label{fig:bandits}Bandit algorithms are not provided any observations (except in contextual bandits); only loss $L_t$ resulting from a particular action $\yp_t$. Nevertheless, they can still be considered as a supervised streaming-learning task under training data $(\yp_1,L_1),\ldots,(\yp_t,L_t)$. } 
\end{figure}

	Multi-arm bandits \cite{RLBook2018} in the context of machine learning can be understood as reinforcement learning without any temporal dependence between state-observations $x_t$ and $x_{t+1}$. Contextual bandits \cite{soemers2018adapting,bouneffouf2020survey} do allow for observations of a state, but (unlike the setting of reinforcement learning) due to the lack of state transition and associated temporal dependence, it is not possible for action $\yp_t$ to affect observations $x_{t+1}, x_{t+2}, \ldots$. Like reinforcement learning, there are never $y_t$-label observations, and bandits are also frequently approached as an online learning task. The setting is shown in Fig.~\ref{fig:bandits}, where we only observe the performance of our classifier according to the loss $L_t$ on our predictions $\yp_t$.

\begin{example}[Online advertisement]  
A stream of anonymous users arrive at a website, $t=1,2,\ldots$. One of three possible advertisements, for the same product, can be shown on each page visit. The goal is to maximize the number of user clicks following the advertisement. Here, $\yp_t \in \{1,2,3\}$ is the action, and the empirical effect is $L_t=0$ (user does \emph{not} click) or $L_t=1$ (user clicks). A bandit algorithm generates and learns from this experience $(\yp_t,L_t)$ via the interaction with visiting users. Thus, there is strong motivation to quickly incorporate learned experience (which advertisement is best) while simultaneously ensuring sufficient experience (to be able to draw statistically-valid conclusions about which advertisement is best).
\end{example}

Other areas of application in the context of streams and bandits (including contextual bandits) are scheduling \cite{kanoun2016big} and fraud-detection \cite{soemers2018adapting}; and many others \cite{bouneffouf2020survey}. Again we emphasize that bandits are traditionally approached in online and stream-like conditions, even when the term `data-stream' is not explicitly mentioned.
Precisely due to the online handling in these scenarios, adversarial settings are an important consideration; where the autonomy of the learning algorithm can be taken advantage of by an attacker \cite{bubeck2012best}.

\subsection{Continual and Lifelong Learning}

Continual or lifelong learning is often associated with deep learning \cite{LifelongLearning,kudithipudi2022biological,parisi2019continual} where the goal is to incorporate information for continuously managing model adaptation, limiting the forgetting of old concepts and tasks. Although data-stream learning is essentially continual learning, in some senses the focus of the data-stream literature dealing with other (non-neural) learning algorithms (such as ensembles of incremental decision trees \cite{AdaptiveRF}) is markedly different. Namely, the research directions in traditional data-stream learning generally deal with relatively low-dimensional sensory data, with emphasis on obtaining good predictive performance under restricted computational resources and other constraints, and detecting or dealing with drift via destructive processes (e.g., restarting learning models). Differently, the lifelong and continuous learning research communities are  more focused on learning inspired by human and biological analogies \cite{kudithipudi2022biological}, often on large text or image collections. Due to the inherent size and cost of training models, a major challenge in lifelong machine learning is to avoid forgetting of earlier tasks and concepts. Recurrent concepts in data streams, however, can be explicitly handled to avoid catastrophic forgetting \cite{gama2014recurrent,moreira2018classifying}.

\subsection{Transfer Learning}

Transfer learning \cite{Pan09,sun2018concept}, domain adaptation \cite{David10}, and meta learning \cite{meta_learning} are research settings related to adaptation to concept drift (also called \emph{domain shift} in these research communities). In the mainstream machine learning (deep learning) literature, these approaches are considered because neural architectures are typically computationally or otherwise costly to train anew
and finding effective ways to reuse them is desired.

Similarly to lifelong learning, the focus in these settings are on more unstructured and higher dimensional data than sensory data typically considered in data streams. 

Apart from that, the key difference is
that in data streams the boundaries between concepts are not explicitly delineated and would need to be inferred from the stream, whereas in transfer, meta learning as well as domain adaptation the source and target tasks need to be explicit. As a result, adaptation to new concepts in the data-stream setting can potentially be easier to automate. 
Models in learning over data streams are thus treated in a more expendable way, and are frequently reset rather than explicitly adapted, and  adaptations \cite{HAT} incur more overriding than adaptation.

\subsection{Change detection and modeling}
\label{sec:drift_detection_matter}

Learning to detect, anticipate, and model concept drift is a major task in the context of data streams. However, since a direct expression of each concept, and knowledge of each, is not available, this task is inherently unsupervised learning and thus outside the scope of discussion in this article which focuses on supervised learning tasks. However, we do discuss the use of change-detection algorithms in the context of supervised learning, in Section~\ref{sec:drift_detection_algorithms}.

\section{Data Stream Learning: Algorithmic Strategies}
\label{sec:methodology_and_strategies}

After overviewing the learning tasks related to learning from data streams 
we next survey algorithmic strategies for data streams.

\subsection{Preliminaries: Definitions for learning modalities}

From the data handling perspective several types of learning are possible over streaming data \cite{almeida2023time}; we list them in order of constraint (from most constrained to least constraints):

\begin{definition}[Online learning]
	A model is updated by each training example $(x_{i},y_{i})$ \emph{exactly once}, and precisely when it is made available.

\end{definition}
 
Since in online learning \cite{lobo2020spiking,kolter2005using}
the model is updated at each time step, recursive definitions for online-learning are possible and indeed typical \cite{jordan1998notes}. Online learning is not possible when labels may be missing or delayed.

\begin{definition}[Single-pass learning]
	A model is updated by each training example $(x_{i},y_{i})$ \emph{at most once}; the decision must be made when the example is made available. 
\end{definition}

Examples can be found in \cite{ditzler2015learning,hou2017one}. Learning pairs may be ignored due to resource constraints, because they contain missing values (among other reasons). Recursive definitions are not as common, due to the possibility to ignore instances (which may break the recursion). 

Single-pass learning has by its name the implication that data cannot be passed through the stream multiple times and in this context it may be considered even stricter than online learning; but such delivery modalities go beyond the discussion in this section where we are considering learning modalities. 

\begin{definition}[Incremental learning]
	The model is updated at any point in time $t$, based on any number and combination of training pairs. 
\end{definition}

Differently from online and single-pass learning, a buffer may be used to store training pairs; such that instances are not necessarily processed in the order of arrival; and also, possibly processed more than once (a buffer may be used to recycle instances through the learning algorithm). This can be useful, e.g., in the context of replay methods \cite{zhang2017deeper,van2022three}, to help break temporal dependence. 

This gives rise to the two varieties, we define as follows.

\begin{definition}[Instance-incremental learning]
	The model is updated using at most a single instance per time step; namely \emph{any} instance already received via the stream.
\end{definition}

Thus, any instance implies: not necessarily the mostly recently-received instance, and possibly an instance which has already been used earlier. 

Without storing instances, instance-incremental learning is equivalent to single-pass learning. Even though a single update is carried out per time step, the buffer can be of arbitrary size (dependent on resource constraints). 

\begin{definition}[Batch-incremental learning]
	The model is updated using any number of training examples (referred to as a batch update) per time step. 
\end{definition}

A buffer of historical data is mandatory in this case. The buffer size does not prescribe update frequency. For instance, it is possible to update the model with a batch of the 10 most-recently delivered instances, at each time step. This will result in reusing every instance 10 times. More generally, any subset of training examples received until the current time (including all or none) may be used in a batch update.

\begin{definition}[Batch learning]
	A model is built offline (prior to deployment) on a batch of instances; and then there is no model update to this model of any kind during deployment; although it may be replaced by a newly-trained model. 
\end{definition}

Batch learning in data stream research refers to the traditional supervised machine learning setting where a model is trained prior to making any predictions.

Table~\ref{tab:this} lists the types of learning just defined, and names some specific learning approaches which may belong to these types, along with their associated complexity for an update. In the following subsections we review these types in more detail.

\begin{table}
	\caption{\label{tab:this}As a summary of which types of approaches can be used under which data stream learning settings. In all cases we refer to the vanilla version of the algorithms, or the version which allows each capability. The final column ($O$ wrt $t$) indicates the complexity at time $t$. For batch methods, $O(1)$ is because the computational cost does not change once model-building is complete. An incremental ensemble (incremental learning) inherits the complexity of the selected model class ($\circ$).}
\begin{tabular}{lcccccc}
\toprule
	Approach             & Online         & Single pass    & Instance-incr. & Batch incr. & Batch offline & $O$ wrt $t$ \\
    \midrule                              
	Naive Bayes          & yes            & yes            & yes            & yes         & yes           & $O(1)$ \\
	Hoeffding trees      & no             & no             & yes            & yes         & yes           & $O(\log t)$ \\
	Nearest neighbors    & no             & no             & yes$^\dagger$  & yes         & yes           & $O(t)$ \\
	Neural networks      & yes$^\ddagger$ & yes$^\ddagger$ & yes            & yes         & yes           & $O(1)$ \\
	Incremental ensemble & no             & no             & no             & yes         & yes           & $\circ$ \\
	Batch method         & no             & no             & no             & no          & yes           & $O(1)$ \\
    \bottomrule
\end{tabular}
$^\dagger$ if one considers an instance-increment to a buffer; $^\ddagger$ implying gradient descent. 
    
\end{table}

Next we survey the main classes of learning-algorithms for streams are within different incremental learning categories. 

\subsection{Instance-incremental learning}
\label{sec:methods}

Recalling that instance-incremental learning encompasses both online and single-pass approaches, we consider the following model classes.

\subsubsection{Generative and Bayesian models} Naive Bayes is the typical baseline for generative classifiers; it makes the assumption of independent features which is both convenient for efficient updates and limiting for expressiveness.

Sufficient statistics can be updated incrementally and online.

Modifications such as tree-augmented naive Bayes (TAN) structures \cite{singharoy2018tree}, that do not rely on the independence assumption, can be used. A related class of models are those carrying out Bayesian inference (e.g., \cite{silva2024streaming}) where, also, probabilities can be updated in an elegant online fashion. Updating probabilities is in a sense simply counting; and sketch approximations to counting from streams can be employed where data dimensions greatly exceed resources \cite{gomes2019machine,MOAbook}. 

\subsubsection{Incremental decision trees} Such methods are popular for data stream learning, including the Hoeffding tree \cite{HolmesKP05}. Arguably, their popularity owes mostly to the abstract attractiveness of an incremental decision tree, coupled with their renown performance in ensembles, in which they almost ubiquitously appear \cite{gomes2017survey}. Like naive Bayes, they are based on counters, and in that sense they are online learners, but only in this sense; even though each instance may be inspected, the actual tree-increments are batch-incremental often with a fixed grace period of non-growth, and the resulting learning algorithm is typically extremely conservative and reluctant to grow new branches even after seeing many instances \cite{Provocative2}. This makes sense because trees cannot be adapted easily once grown, and even if methods do certainly exist to adapt them \cite{HAT,ExtremelyFastDecisionTrees}, the most popular approaches to adaptation are still the most destructive. It is common to put other (e.g., online-learning) algorithms at the leaves of a Hoeffding tree; with naive Bayes being an obvious choice because it uses the same statistics gathered.

\subsubsection{Instance-based learning and nearest neighbor methods} Methods like $k$-nearest neighbors ($k$NN) are also heavily favored in streams \cite{pmlr-v94-roseberry18a,roseberry2021self,melgar2021nearest}, as they are naturally suited to dealing with concept drift in the sense that a natural forgetting (and thus, adaptation) mechanism is forced upon the model by the simple infeasibility of storing and searching through an ever-growing buffer of instances. Variably-sized buffers are typical, and external concept-drift detection mechanisms can be incorporated anyway to flush old concepts (instances) from the buffer more quickly. As a non-parametric method, $k$NN is not online-learning in the parametric sense, but its decision function may be updated one instance at a time (unlike a naive implementation of Hoeffding tree induction). 

\subsubsection{Neural networks via stochastic optimization} Such an approach is inherently instance-incremental, since learning can be continued from a model at any point and can incorporate new instances. 
This category includes any stochastic gradient-descent-based methods, including logistic regression.
Catastrophic forgetting \cite{kirkpatrick2017overcoming} is an associated term that refers to putting too much emphasis on the most recent learning iterations.
Mini-batch increments are typical \cite{DeepLearningBook}. 

\subsection{Batch-incremental learning}
\label{sec:modeling}

Ensembles methods fit elegantly into this category; having multiple models trained on the same task, but on different selections of data, can mitigate risks and uncertainties, for example with regard to drift detection (when to reset a model) and hyper-parametrization (how large a batch size, etc.) \cite{Kuncheva_book,Zliobaite_thesis,gomes2017survey}.
Decision trees prove popular \cite{kolter2005using,ghazikhani2013ensemble,AdaptiveRF}, but any type of model may be considered. Note that ensemble members can be either incremental, or batch-offline learners. Even if individual models were trained in a traditional batch setting, the way that they are phased in or out constitutes a batch-incremental approach. Neural networks (not necessarily ensembles) also fit naturally into this area; as just mentioned above.

\subsection{Batch learning}
\label{sec:batch_learning}

The traditional machine learning approach of learning from static batches of data applies also to the streaming scenario. Here, a large batch of data is collected from the stream; a model is then trained and deployed for prediction on instances (say, at time $t$), then (at $t+1,t+2,\ldots,\infty$) predictions are made as required, but no more learning is carried out. The model may be replaced, but unlike batch-incremental learning, learning is not performed live on the stream.

\subsection{On the use of drift-detection algorithms}
\label{sec:drift_detection_algorithms}

Concept-drift (or change-point) detection mechanisms (e.g., \cite{ADWIN,CUMSUM}) are often incorporated by a variety of data-stream learning strategies. The general motivation is to mitigate performance loss when there is a change in concept, by responding rapidly to refresh or replace existing models \cite{sebastiao2009study}. For example, in \textit{destructive adaptation}, a number of decision trees in a random forest might be reset \cite{AdaptiveRF}, or the buffer of k-nearest neighbor could be emptied \cite{roseberry2021self}. In \textit{continuous adaptation}, the learning rate might be increased for stochastic-gradient descent \cite{Provocative1,ghazikhani2013ensemble}.  

Authors typically use model parameters or error (denoted elsewhere in this work as $\theta_t$ and $L_t$, respectively) streams as a proxy to speculate on the character and movement of a concept \cite{you2021learning,Gama04,ADWIN,ESR}, and some authors model concept $P_t$ explicitly \cite{Hanen2015,haug2021learning}.

If concept drift is uncertain, or determined to be of small magnitude, explicit adaptation may not be preferred. An alternative strategy is to build robustness into a model such that it may remain in use without change, after drift.

\subsection{Strategies for handling temporal dependence}
\label{sec:temporal_dep}

The assumption of drift already implies some form of temporal dependence in the context of a stream \cite{Provocative1}, however, temporal dependence may also be exhibited within a single concept, including the tasks of filtering, forecasting, smoothing; as discussed above (recall, e.g., Table~\ref{tab:time_series}). There are two main approaches for appropriately handling this dependence: 
\begin{enumerate}
	\item A moving window, as described variously above, such as in Example~\ref{ex:temp} and Section~\ref{sec:temp_sec}; which is equivalent to Markovian state space models \cite{Our2015Paper,SequentialDataReview} and time-delay neural networks \cite{DuSwamy}; or other kinds of feature engineering and preprocessing
	\item A (typically, neural-) representation including recurrent, convolution, and attention mechanisms \cite{DuSwamy,iqbal2024anomaly,lim2021time,chen2021developing,lim2021time}
\end{enumerate}
or some combination of the above. An in-depth coverage is application dependent and beyond the scope of this article; we direct the reader to the references and references therein.

\section{Recommendations: research meets practice}

Our updates and synthesis in this paper are aimed to clear up confusion that originated in the stream literature when moving from data-stream processing (treat each item in a stream at speed) to learning. Under learning, the constraints of item-based processing urgency is typically superseded by the importance of minimizing a loss metric, where doing so robustly in order to generalise to new data and doing so with high accuracy is more important than doing so quickly. Even if the time-granularity of delivery is in milliseconds (per instance arrival), model updates may be better carried out every day, week, or even be spaced many months apart. 

Data-streams research has come a long way. We provide a historical perspective in \cite{HistoricalPerspective}, on the origin of the term and the constraints frequently imposed by data stream processing. While a myriad of techniques for different tasks and settings is already available, not all of them have practical relevance beyond academic curiosity.  In order to get a clearer idea of needs and constraints as they appear in practice, and match algorithmic strategies available in the academic literature to different problem settings and challenges for deployment, we conducted an informal survey among several past and present industrial collaborators and research practitioners who encounter data streams in the context of which they wish to deploy supervised learning algorithms. Our respondents represented the sectors of banking, finance, transportation, energy, and medicine. While these interviews were informal and unstructured, our questions stemmed from various assumptions, constraints, and other \emph{communis opinio} that we found in the academic literature on data streams; outlined in the subsections as follows.

\subsection{How soon after a data observation arrives is a corresponding prediction required?}

The assumption that a model is immediately ready to provide a prediction is often made in the data streams literature (e.g., \cite{zhai2017classification,MOAbook}), usually on the basis that data is arriving quickly; but this context is not specific to learning from data streams. 

When provided with an input instance, any classification or regression model can provide a prediction. It is true that some methods may take longer than others in their inference process, but any time restrictions are not inherent to data streams, but more related to the application domain (e.g., for autonomous driving or  financial trading reaction times in milliseconds may be needed).

The requirement to predict immediately can also be made on the assumption that batch-learning models may still be in training, not ready for deployment, with training data still being gathered. One can simply output a default prediction or use a basic model until enough data has arrived to train the batch learner. But it is worth reflecting on what  practical applications one would realistically need to or wish to deploy a stream-learning algorithm with no prior data, no prior knowledge about its prediction task, and no existing model. 

It can be that even when a model is ready to provide a prediction the instance may not yet be available. For example, remote sensor observations can be delayed.

As we have mentioned with regard to the smoothing task: a model be ready to supply predictions only when required, not earlier, as it might imply a penalty in terms of predictive performance. 

\subsection{When and how frequently do labels become available for model training?} 

Although not ubiquitous, an assumption prescribing that a stream of training labels is immediately available to the learner is relatively common in research on stream learning. The only mechanism for providing training labels immediately, rapidly, and fully reliably, is where future forecasting targets become known via the passing of time. In such a case, most likely the underlying data is of a time series nature exhibiting temporally-dependencies \cite{Provocative2}.
If a rapid and reliable process already exists with which to obtain ground-truth labels for learning, then likely there is no need for immediate or online learning. If the process exists only partially (e.g., delayed \cite{grzenda2020delayed} or sparse \cite{Fahy23} labeling) then classical online learning is not possible. Hence we argue that using an online learning paradigm is only futile for real data streams, except in forecasting tasks when the aim is to predict outcomes that will happen in the future. 

\subsection{Is an online model update required with each new training example?} 
\label{sec:instantaneous}

Single-pass online learning should only be considered when it is practically possible (i.e., when labels are reliably available without delay). Since this case is not always realizable, it is encouraging that settings of delayed or partial labels are indeed being increasingly recognized as more practically-relevant \cite{Plasse16,SemiSupDelayedSurvey,gomes2019machine}, as well as active learning \cite{Active}. 

Even when online learning is feasible, one should carefully consider the risks of this level of instance-to-instance learning autonomy, the limits in accuracy, and the computational cost (online learning can be fast per model-update, but larger less frequent updates can be faster). 

The community often cites the need for learning autonomy (e.g., instance-incremental updates) with adapting to drift (restarting models). However, if there are frequent major and abrupt shifts in concept, we would question if there is not some underlying deployment problem that might put into question a data-stream learning process (and instead suggest more robustness verification, or human intervention). 

Indeed, online learning may be dangerous in safety-critical applications where interference (including adversarial \cite{AdversarialOnline}, but not limited to) may occur. Forcing a learning update at each step regardless of stream behavior raises serious questions of robustness, reliability, and interpretation \cite{haug2022dynamic}; and consideration of the possibilities and consequences that the model rapidly deteriorates; and to promote swift recovery from such an event.
Offline batch learning is thus heavily favored in industrial settings. 

Case in point: chatbots and large language models (LLMs). In 2017, Microsoft's chatbot Tay had to be shut down only hours after launch due to its online algorithm rapidly assimilating and replicating the inappropriate commentary of its users \cite{wolf2017we}. Modern LLMs such as ChatGPT are trained on static datasets with a definite cutoff ($t$) \cite{openai_gpt_2024}, i.e., batch-offline training. The model parameters fixed during training are not updated automatically during its operation (deployment); and updates imply a new version release (batch update). Beyond reasons of model stability and safety, it is compatible with the common approach to building deep neural architectures (cloud-based rather than stream-based data sources).

To be reliable in the long term, it must be assumed that concept drift is relatively rare, minor in extent, or extremely gradual. In practice, these may frequently be valid assumptions, even in relatively dynamic settings. 

And, if there are no major changes to the concept, we can say there is increasingly less need for timely updates as more data is received. Simply: the $t=1\,000\,000$-th training example constitutes roughly one millionth of the knowledge seen so far and is unlikely to have major ramifications on the decision surface (if regression) or decision boundary (if classification). If it were to, then it is more likely a serious anomaly. One could consider decaying or cost-sensitive need for updates \cite{vzliobaite2015towards}.

Thus we propose that online or instance-incremental learning should not exist as a constraint, except out of academic curiosity, or in the case of empirically-demonstrated effectiveness against the more-traditional alternatives. Rather, efforts should be taken to ensure robustness, trustworthiness, and reliable evaluation and monitoring. The testing phase should be close to real operating conditions. 

Finally, while some claim that incremental learning is the ultimate goal of AI \cite{luo2020appraisal}, we disagree, at least in the strict sense; the biological underpinnings of lifelong learning machines are complex \cite{kudithipudi2022biological}, uncertain and suggest a diversity of mechanisms. In the world of data-stream learning, we argue for at least a mix between instance-incremental and batch-incremental (e.g., by \cite{FastAndSlow}), if not more of the latter.

\subsection{Is it possible to store more than one data observation in the operating memory?}

In some of the literature, the use of a buffer for the storage of instances and / or processing is not taken as a principle related to streams \cite{Domingos03,aggarwal2009data,ksieniewicz2019data}.

However, aside from academic curiosity, there is no need to self-impose such artificial constraints. If storage is available to the learning system, maintaining and using a buffer of training instances can lead to better performance than online learning. And model updates may even be more efficient since, if there is a buffer, updates can be less frequent.

In any case, a buffer may be required as an integral part of the model such as in k-nearest neighbors approaches \cite{roseberry2021self,IDA2012} and experience-replay methods \cite{zhang2017deeper,van2022three}. Contemporary learned models often require much more memory than a single learning instance. 
Thus, even the smallest of embedded devices containing machine-learned models are likely to also have memory to hold at least a few instances.

Only in very rare cases, a single data observation or single training instance would overwhelm the memory of a device. This should be declared as a special case, stated as a constraint, and studied specifically.

Storing personal data is becoming increasingly restrictive (for example, the General Data Protection Regulation law in Europe), but even in such cases storing for analytical purposes is usually permissive if collecting is permissive. When data must be protected for privacy reasons, special areas of research, such as federated learning \cite{marfoq2023federated,mawuli2023semi}, can provide alternative solutions.

Other than data protection, one may wish to impose certain constraints on resource use for efficiency or frugality. An often-stated assumption in data-stream learning and predictive systems is that methods should operate under constant computational resources 
(\cite{gepperth2016incremental}). This assumption is relevant in all practical machine learning, computer science, and beyond, not only data streams. The amount of permissible resources is inherently application dependent, yet it can never be unlimited theoretically.

Practically, however, nearly unlimited resources can be accessible, even if costly. For example, in cloud computing, one can acquire additional resources on-the-fly; this can apply to learning from data streams as well. Thus, instead of imposing constant resources per instance, it may be more conceivable to ask what would be the most cost-efficient way of learning \cite{vzliobaite2015towards}.

Energy efficient computing is attracting attention in modern machine learning, including streams \cite{garcia2020energy}, and should be taken forward. However, efficiency does not imply constant resources per instance processed. Often savings may be achieved via tradeoffs where processing some instances takes more resources while savings are achieved on others.

\subsection{Is it necessary to detect concept drift?}
\label{sec:know_drift}

Supervised adaptive learning over data streams can work or without change detection \cite{IndresSurvey}.
Although drift-detection mechanisms often preferred, we make the following observations regarding their use in data-stream learning for building supervised models:
\begin{itemize}
	\item The accuracy of such mechanisms has been infrequently assessed in real-world streams, and naturally so due to the subjective nature of what constitutes a concept or a concept drift. False drift alarms can deteriorate the predictive accuracy of the learned models \cite{Our2015Paper}.
	\item It is not possible to distinguish between a single point anomaly and change-point at current time $t$; meaning that drift detection is always either retrospective or inaccurate. 
    \item Both false negatives (late or missed detections) and false positives (false alarms) can reduce accuracy compared to periodic updates or no updates. 
	\item Even if drift were accurately detected, such detections are not necessarily useful, as drift does not necessarily cause the error rate to rise; the error may be unaffected or even decrease under drift \cite{Zliobaite_thesis,Our2015Paper}. At most, drift rate and error rate may be stochastically correlated \cite{Kuncheva09}. Expunging models after a minor or prolonged ongoing drift can lead to significantly worse performance than ignoring the drift \cite{Provocative2}. 
	\item Detection (and alerts) of possible drift may be insufficient, without also quantifying the magnitude, nature, and extent (in time) of drift. 

\end{itemize}

Thus, it is not mandatory for adaptive learning to deploy drift-detection mechanisms, and even in case of such deployment it is not necessary that these mechanisms be accurate. Drift detection usually implies model refreshment \cite{sebastiao2009study}, and this in turn may offer a forgetting mechanism, increased diversity among ensemble members, better generalization through a simpler model and thus less overfitting, or better learning of the concept in course.

Therefore, if concept-drift detection and modeling is to be considered, we recommend careful study of the effectiveness of detection on the predictive accuracy of the model, under different hypotheses, and against a variety of baselines (including no drift, random drift, simplistic vs complex methods, etc).

\subsection{When and how to update the model?}
\label{sec:when_to_update_a_model}

Even if drift detection is not an objective in itself it is still reasonable to consider a possibility that drift will occur and the model may need to be updated during the time of operation.

There is a trade-off between data horizon (how quickly the most recent data point should be incorporated into a model), and data obsolescence (how long it takes for a data point to become irrelevant to the model, if ever) \cite{Domingos03}. 

In practice, the user inevitably needs to gather some prior-knowledge about the speed and type of drift to be expected. 
Choosing and configuring an algorithm that should learn and perform autonomously in a scenario of completely unknown dynamics is  infeasible in a similar way as there is ``no free lunch" of classical machine learning \cite{Wolpert97}.

The scenario of a short horizon (rapid concept drift) often but not always implies quicker obsolescence (a concept remains relevant for a short time). The relation depends on the magnitude of drift \cite{Zliobaite_thesis}. When drifts are severe such that the two concepts are unrelated, change detection in combination with online learning may be useful. And, in the opposing case of a stable concept, a slower learning strategy is likely to be the better strategy, especially if the concept is complex and thus requires a complex model to represent it. 

Although inherently wasteful of computing resources, destructive adaptation can be effective, especially when we expect many abrupt changes to concepts in the stream. It is much less desired when we expect complex and stable or slowly-incrementing concepts. In that case, continuous adaptation may be preferred, although in practice such approaches will require more monitoring and careful dynamic and automated self-tuning \cite{rauba2024self}.

There are risks with any kind of autonomous online or incremental learning, especially with the inherently difficulty of predicting drift. Algorithms might self-destruct under false (or, real) drift alerts, and continuously-adapting methods may phase out too much knowledge. It can therefore be a sensible option to rescind any type of adaptation, and instead place the focus on anomaly detection and alerts for human intervention; and/or on general robustness to concept drift. In general, interpretation and explainability are crucial aspects to consider not only behind making predictions but also behind change detection and adaptation mechanisms.

\subsubsection{The case for and against incremental decision trees}

The Hoeffding tree learning algorithm deserves special consideration, given that methods derived from Hoeffding trees, including ensembles of such trees, have often been shown top accuracy in state-of-the-art data-stream comparisons \cite{AdaptiveRF}. 
Despite Hoeffding trees having risen to prominence in data-stream learning research, they are empirically disadvantageous, at least as standalone models \cite{IDA2012,Provocative2,RandomSubSpaces} as they are slow to grow and inflexible once grown. In addition, they have been shown to be mathematically unjustified for the stream settings  \cite{matuszyk2013correcting,rutkowski2013decision}.

Hoeffding trees can be considered stream learning only under the loosest assumptions of what is a data stream. With only some internal counters being incremented per training example, with very infrequent changes to the tree structure itself, which is indeed slow to grow and even explicitly prevented during a grace period of typically hundreds of instances while tree-growth is explicitly inhibited. This makes sense, since the price to pay is increasing complexity in relation to the number of instances processed over time. Thus more learning corresponds to more time-consuming prediction, as instances need to be propagated down to an ever-deeper tree. They are inherently difficult to adapt to evolving concepts and, while some approximations exist \cite{HAT,ExtremelyFastDecisionTrees}, they require additional components like change detectors that add to computational costs. 

As a result, in the academic setting, truly instance-incremental methods like naive Bayes, stochastic gradient descent, or even Kalman filters are typically used at the leaves \cite{ziffer2021kalman}, and they are deployed in large dynamic ensembles under constant destruction and renewal in order to exhibit robustness against drift. It is the ensembles and the models at the leaves that probably should carry most of the predictive power attributed to Hoeffding trees.

A dynamic ensemble also provides an inherent adaptability that Hoeffding trees need to operate in the data stream setting. Without it, the problem of catastrophic \emph{remembering}, where model parts built on old concepts are forever carried forward, becomes an issue. We propose that Hoeffding trees should only be considered for data that has a small number of features
with relatively simple concepts, while images and natural language should be handed over to the neural-network framework, or batch methods should be considered instead.

\subsection{Does the data include temporal dependencies?}
\label{sec:data}

The existence of temporal dependence in data streams 
occurs widely in practice, but temporal dependence is a widely studied phenomenon; we provided some references in Section~\ref{sec:temporal_dep}.

Lack of realism in the testing scenarios has been one of the main criticisms of data streams research over the years. A popular benchmark dataset \textsf{Electricity} \cite{Electricity} is still in regular use in empirical evaluations of the data-stream learning literature despite long standing criticisms \cite{Zliobaite13electricity}. 

The source paper \cite{Electricity0} does not mention streams, does not use online or instance-incremental learners, and states the main goal as an investigation of identifying and `exploiting hidden contexts' but does not point to any drift. This data set has been constructed as an academic exercise and is disconnected from practical challenges. Having a model to match supply and demand in real time is actually an essential problem facing energy suppliers \cite{suganthi2012energy,fekri2021deep,melgar2021nearest}, especially with the rise of renewable energy sources. However, learning such a model in real-time from a concept-drifting data stream without human expertise or intervention, is unrealistic.

Other commonly used datasets including \textsf{CoverType} or \textsf{PokerHand} have nothing to do with streams. Drifts and temporal dependence in those datasets come from artificial ordering of instances, with no basis we of aware of on challenges that exist in practice. Calling these benchmarks real-world datasets should be avoided.

\subsection{Is the data high-dimensional?}

Beyond the temporal realism of concepts and drift, there is the question of size; including the dimensionality of instances. 

In some of the traditional data-stream literature, even less than 100 features may be considered high-dimensional data \cite{aggarwal2004framework,zhai2017classification,AdaptiveRF,das2019learn}. In this context, consistently restarting the learning process to ensure adaptation to the current concept may be feasible, since high-performing models can be learned relatively rapidly from training examples of compact feature representations. 

On the other hand, high-dimensional data usually implies complex concepts requiring high-capacity models. Therefore, we see a trend of almost exclusively neural-network architectures for data streams of instances  defined by many thousands of features; particularly collections of unstructured (non-tabular) image or textual representations \cite{pinckaers2020streaming,maggini2019learning}. This makes sense, as deep-learning methodologies are not only well-suited to learning from large datasets \cite{DeepLearningBook} but, in a stream setting, online or incremental updates are natural \cite{lobo2020spiking} and such learning updates can be constant over time (cf.~Table~\ref{tab:this}).

Thus we observe that recommendations among learning strategies is heavily driven by questions on the dimensionality of data; significantly more so than questions of stream speed or concept drift. This appears to have led to the formation of two distinct stream communities; one based on stream-learning frameworks showcasing mainly ensembles of incremental decision trees (\cite{MOAbook,CapyMOA}), and one considering relatively large and complex neural architectures often under the umbrella of continual or lifelong learning.

Learning complex concepts from high-dimensional data is a costly investment, and thus, avoiding forgetting concepts already learned becomes a priority, much conversely to simplistic low-dimensional concepts. For large models such as LLMs of millions or billions of parameters, quickly transitioning from one concept to another is out of the question; and cautious batch updates, modeling many sub-concepts simultaneously, make more sense, as discussed above in Section~\ref{sec:instantaneous}.

We argue that online and incremental learning methods are better suited closer to the edge where data is likely to be smaller (in terms of features), faster, with more drift among concepts. But it is essential to perform a thorough assessment of these aspects alongside the actual performance needs and constraints of a deployment, during task formulation. 

Conversely, for complex concepts represented by high-dimensional data, we recommend moving learning processes toward the cloud wherever possible, where models can be trained via incremental or batch (version) updates, better leveraging the high capacity of neural architectures to learn multiple tasks and concepts from large data lakes. There, these models can be heavily tested before being deployed in a streaming setting, if such a streaming deployment is indeed necessary. Here, robustness to out-of-domain samples, anomalies, and other volatility remains a priority over autonomous adaptation, and greater emphasis may be placed on data privacy, especially since large models have more capacity to memorize information. Regularizing techniques and differential privacy methods can help mitigate these risks while ensuring performance at scale. An in-depth discussion on privacy of large models is beyond the scope of this paper; see, e.g., \cite{das2025security,yao2024survey}.

\section{Concluding Remarks}
\label{sec:conclusions}

Machine learning from data streams continues to be a popular and practically relevant research area, that has attracted major attention from the scientific community. Here we have provided a research synthesis of supervised learning from data streams and discussed them in the context of potential practical applications.

We propose that many assumptions in the data streams literature do not match to supervised-learning task settings that corresponds to a practical application. Many practical deployments do not involve any stream-specific assumptions or constraints typically assumed in the academic literature, meaning that the traditional batch-based learning can be employed. Specifically, we suggest that single-pass and online-learning constraints be decoupled from the data-stream learning methodology. In addition, instead of focusing on scenarios that emphasize frequent large sudden drifts, we would recommend placing more attention towards surveying and characterizing how, how-often, and what kind of drifts happen can be realistically expected. 

We also suggest to turn more attention of data-stream research to the robustness and trustworthiness of models that can be trained offline and deployed on streaming data. That is, how to build relatively static models which are robust to drift, and can continue to perform autonomously without restarting any learning process. This could also lead to a more effective interchange with the continual and transfer learning research communities. 

Learning tasks beyond that of classification have been largely overlooked in the data streams literature. Arguably, the regression tasks of filtering and forecasting are more naturally suited to stream settings than classification. 

Indeed, we find that the tasks most closely matching the traditional data-stream learning settings, involve  interacting in a dynamic environment rather than on static data sets; for example imitation learning, reinforcement learning, and bandits. Meanwhile, neural transfer learning provides natural increment-ability and transferability to new concepts, as well as mechanisms to adapt quickly without destructive adaptation practices requiring to restart models from scratch, for instance, by using variable learning rates. Traditional machine learning approaches, such as decision trees and their ensemble variants, may still hold their ground on relatively low-dimensional tabular data, but we doubt the necessity and practical benefits of enforcing the use of online or incremental varieties.

The deployment of online or incremental learning algorithms is autonomous learning, and carries major risks. It is not surprising that many practitioners prefer instead a traditional segmented approach of training, validation and testing, and then deployment. We recommend a shift of research focus from stream-learning and adaptation to concept drift under constraints, to robustness, interpretability, and safety of models.

\bibliographystyle{plain}
\bibliography{references}

\begin{thebibliography}{100}

\bibitem{Aggarwal_book}
Charu Aggarwal, editor.
\newblock {\em Data Streams Models and Algorithms}.
\newblock Springer, 2007.

\bibitem{aggarwal2009data}
Charu Aggarwal.
\newblock Data streams: An overview and scientific applications.
\newblock {\em Scientific Data Mining and Knowledge Discovery: Principles and
  Foundations}, pages 377--397, 2009.

\bibitem{aggarwal2004framework}
Charu Aggarwal, Jiawei Han, Jianyong Wang, and Philip Yu.
\newblock A framework for projected clustering of high dimensional data
  streams.
\newblock In {\em Proceedings of the Thirtieth international conference on Very
  large data bases-Volume 30}, pages 852--863, 2004.

\bibitem{almeida2023time}
Ana Almeida, Susana Br{\'a}s, Susana Sargento, and Filipe~Cabral Pinto.
\newblock Time series big data: a survey on data stream frameworks, analysis
  and algorithms.
\newblock {\em Journal of Big Data}, 10(1):83, 2023.

\bibitem{alzghoul2012data}
Ahmad Alzghoul, Magnus L{\"o}fstrand, and Bj{\"o}rn Backe.
\newblock Data stream forecasting for system fault prediction.
\newblock {\em Computers \& industrial engineering}, 62(4):972--978, 2012.

\bibitem{Barber}
David Barber.
\newblock {\em {Bayesian Reasoning and Machine Learning}}.
\newblock {Cambridge University Press}, 2012.

\bibitem{Basseville_Nikoforov}
Michele Basseville and Igor Nikiforov.
\newblock {\em Detection of Abrupt Changes: Theory and Application}.
\newblock Prentice-Hall, 1993.

\bibitem{David10}
Shai Ben-David, John Blitzer, Koby Crammer, Alex Kulesza, Fernando Pereira, and
  Jennifer Vaughan.
\newblock A theory of learning from different domains.
\newblock {\em Mach. Learn.}, 79(1–2):151–175, 2010.

\bibitem{ADWIN}
A.~Bifet and R.~Gavalda.
\newblock Learning from time-changing data with adaptive windowing.
\newblock In {\em Proc. of the Seventh SIAM Int. Conf. on Data Mining}, 2007.

\bibitem{HAT}
Albert Bifet and Ricard Gavalda.
\newblock Adaptive learning from evolving data streams.
\newblock In {\em IDA 8th Int. Symposium on Intelligent Data Analysis}, pages
  249--260, 2009.

\bibitem{MOAbook}
Albert Bifet, Ricard Gavald\`a, Geoff Holmes, and Bernhard Pfahringer.
\newblock {\em Machine Learning for Data Streams with Practical Examples in
  MOA}.
\newblock MIT Press, 2018.
\newblock \url{https://moa.cms.waikato.ac.nz/book/}.

\bibitem{Hanen2015}
Hanen Borchani, Ana~M. Mart{\'{\i}}nez, Andr{\'{e}}s~R. Masegosa, Helge
  Langseth, Thomas~D. Nielsen, Antonio Salmer{\'{o}}n, Antonio Fern{\'{a}}ndez,
  Anders~L. Madsen, and Ram{\'{o}}n S{\'{a}}ez.
\newblock Modeling concept drift: {A} probabilistic graphical model based
  approach.
\newblock In {\em IDA 2015: 14th International Symposium on Intelligent Data
  Analysis}, pages 72--83, 2015.

\bibitem{bouneffouf2020survey}
Djallel Bouneffouf, Irina Rish, and Charu Aggarwal.
\newblock Survey on applications of multi-armed and contextual bandits.
\newblock In {\em 2020 IEEE congress on evolutionary computation (CEC)}, pages
  1--8. IEEE, 2020.

\bibitem{bubeck2012best}
S{\'e}bastien Bubeck and Aleksandrs Slivkins.
\newblock The best of both worlds: Stochastic and adversarial bandits.
\newblock In {\em Conference on Learning Theory}, pages 42--1. JMLR Workshop
  and Conference Proceedings, 2012.

\bibitem{chang2025model}
Yingshan Chang and Yonatan Bisk.
\newblock Model successor functions.
\newblock {\em arXiv preprint arXiv:2502.00197}, 2025.

\bibitem{chen2021developing}
Xie Chen, Yu~Wu, Zhenghao Wang, Shujie Liu, and Jinyu Li.
\newblock Developing real-time streaming transformer transducer for speech
  recognition on large-scale dataset.
\newblock In {\em ICASSP 2021-2021 IEEE International Conference on Acoustics,
  Speech and Signal Processing (ICASSP)}, pages 5904--5908. IEEE, 2021.

\bibitem{LifelongLearning}
Zhiyuan Chen and Bing Liu.
\newblock Lifelong machine learning.
\newblock {\em Synthesis Lectures on Artificial Intelligence and Machine
  Learning}, 12(3):1--207, 2018.

\bibitem{coraddu2024floating}
Andrea Coraddu, Luca Oneto, Jake Walker, Katarzyna Patryniak, Arran Prothero,
  and Maurizio Collu.
\newblock Floating offshore wind turbine mooring line sections health status
  nowcasting: From supervised shallow to weakly supervised deep learning.
\newblock {\em Mechanical Systems and Signal Processing}, 216:111446, 2024.

\bibitem{Dada19}
Emmanuel~Gbenga Dada, Joseph~Stephen Bassi, Haruna Chiroma, Adebayo~Olusola
  Adetunmbi, Opeyemi~Emmanuel Ajibuwa, et~al.
\newblock Machine learning for email spam filtering: review, approaches and
  open research problems.
\newblock {\em Heliyon}, 5(6):e01802, 2019.

\bibitem{das2019learn}
Ariyam Das, Jin Wang, Sahil~M Gandhi, Jae Lee, Wei Wang, and Carlo Zaniolo.
\newblock Learn smart with less: Building better online decision trees with
  fewer training examples.
\newblock In {\em IJCAI}, pages 2209--2215, 2019.

\bibitem{das2025security}
Badhan~Chandra Das, M~Hadi Amini, and Yanzhao Wu.
\newblock Security and privacy challenges of large language models: A survey.
\newblock {\em ACM Computing Surveys}, 57(6):1--39, 2025.

\bibitem{Mathelin25}
Antoine de~Mathelin, Fran{\c{c}}ois Deheeger, Mathilde Mougeot, and Nicolas
  Vayatis.
\newblock Deep out-of-distribution uncertainty quantification via weight
  entropy maximization.
\newblock {\em Journal of Machine Learning Research}, 26(4):1--68, 2025.

\bibitem{desai2019real}
Angel~N Desai, Moritz~UG Kraemer, Sangeeta Bhatia, Anne Cori, Pierre Nouvellet,
  Mark Herringer, Emily~L Cohn, Malwina Carrion, John~S Brownstein, Lawrence~C
  Madoff, et~al.
\newblock Real-time epidemic forecasting: challenges and opportunities.
\newblock {\em Health security}, 17(4):268--275, 2019.

\bibitem{SequentialDataReview}
Thomas~G. Dietterich.
\newblock Machine learning for sequential data: A review.
\newblock In {\em Proceedings of the Joint IAPR International Workshop on
  Structural, Syntactic, and Statistical Pattern Recognition}, pages 15--30,
  London, U.K., 2002. Springer-Verlag.

\bibitem{Disabato22}
S.~Disabato and M.~Roveri.
\newblock Tiny machine learning for concept drift.
\newblock {\em IEEE Transactions on Neural Networks and Learning Systems},
  2022.

\bibitem{ditzler2015learning}
Gregory Ditzler, Manuel Roveri, Cesare Alippi, and Robi Polikar.
\newblock Learning in nonstationary environments: A survey.
\newblock {\em IEEE Comp. Intell. Magazine}, 10(4):12--25, 2015.

\bibitem{Domingos03}
Pedro Domingos and Geoff Hulten.
\newblock A general framework for mining massive data streams.
\newblock {\em Journal of Computational and Graphical Statistics}, 12:945--949,
  2003.

\bibitem{DuSwamy}
Ke-Lin Du and M.~N.S. Swamy.
\newblock {\em Neural Networks and Statistical Learning}.
\newblock Springer Publishing Company, Incorporated, 2013.

\bibitem{Fahy23}
Conor Fahy, Shengxiang Yang, and Mario Gongora.
\newblock Scarcity of labels in non-stationary data streams: A survey.
\newblock {\em ACM Computing Surveys}, 55(2):1--39, 2023.

\bibitem{fekri2021deep}
Mohammad~Navid Fekri, Harsh Patel, Katarina Grolinger, and Vinay Sharma.
\newblock Deep learning for load forecasting with smart meter data: Online
  adaptive recurrent neural network.
\newblock {\em Applied Energy}, 282:116177, 2021.

\bibitem{Gama04}
J.~Gama, P.~Medas, G.~Castillo, and P.~Rodrigues.
\newblock Learning with drift detection.
\newblock In {\em Advances in Artificial Intelligence}, pages 286--295, 2004.

\bibitem{Gama_book}
Joao Gama.
\newblock {\em Knowledge Discovery from Data Streams}.
\newblock Chapman \& Hall, 2010.

\bibitem{meta_learning}
Joao Gama and Peter Kosina.
\newblock Learning about the learning process.
\newblock In {\em Advances in Intelligent Data Analysis X}, pages 162--172,
  2011.

\bibitem{gama2014recurrent}
Joao Gama and Petr Kosina.
\newblock Recurrent concepts in data streams classification.
\newblock {\em Knowledge and Information Systems}, 40:489--507, 2014.

\bibitem{Electricity}
Joao Gama, Pedro Medas, Gladys Castillo, and Pedro Rodrigues.
\newblock Learning with drift detection.
\newblock In {\em Brazilian symposium on artificial intelligence}, pages
  286--295, 2004.

\bibitem{IndresSurvey}
Joao Gama, Indre Zliobaite, Albert Bifet, Mykola Pechenizkiy, and Abdelhamid
  Bouchachia.
\newblock A survey on concept drift adaptation.
\newblock {\em {ACM} Computing Surveys}, 46(4):44:1--44:37, 2014.

\bibitem{garcia2020energy}
Eva Garc{\'\i}a~Mart{\'\i}n.
\newblock {\em Energy Efficiency in Machine Learning: Approaches to Sustainable
  Data Stream Mining}.
\newblock PhD thesis, Blekinge Tekniska H{\"o}gskola, 2020.

\bibitem{gardner2023benchmarking}
Josh Gardner, Zoran Popovic, and Ludwig Schmidt.
\newblock Benchmarking distribution shift in tabular data with tableshift.
\newblock {\em Advances in Neural Information Processing Systems},
  36:53385--53432, 2023.

\bibitem{gepperth2016incremental}
Alexander Gepperth and Barbara Hammer.
\newblock Incremental learning algorithms and applications.
\newblock In {\em European symposium on artificial neural networks (ESANN)},
  2016.

\bibitem{ghazikhani2013ensemble}
Adel Ghazikhani, Reza Monsefi, and Hadi~Sadoghi Yazdi.
\newblock Ensemble of online neural networks for non-stationary and imbalanced
  data streams.
\newblock {\em Neurocomputing}, 122:535--544, 2013.

\bibitem{gomes2017survey}
Heitor Gomes, Jean~Paul Barddal, Fernanda Enembreck, and Albert Bifet.
\newblock A survey on ensemble learning for data stream classification.
\newblock {\em ACM Computing Surveys (CSUR)}, 50(2):1--36, 2017.

\bibitem{AdaptiveRF}
Heitor Gomes, Albert Bifet, Jesse Read, Jean Barddal, Fabrício Enembreck,
  Bernhard Pfahringer, Geoff Holmes, and Talel Abdessalem.
\newblock Adaptive random forests for evolving data stream classification.
\newblock {\em Machine Learning Journal}, 106(9-10):1469--1495, 2017.

\bibitem{SemiSupDelayedSurvey}
Heitor Gomes, Maciej Grzenda, Rodrigo Mello, Jesse Read, Minh Huong~Le Nguyen,
  and Albert Bifet.
\newblock A survey on semi-supervised learning for delayed partially labelled
  data streams.
\newblock {\em Computing Surveys}, 2022.

\bibitem{CapyMOA}
Heitor Gomes, Anton Lee, Nuwan Gunasekara, Yibin Sun, Guilherme~Weigert
  Cassales, Justin~Jia Liu, Marco Heyden, Vitor Cerqueira, Maroua Bahri,
  Yun~Sing Koh, Bernhard Pfahringer, and Albert Bifet.
\newblock {CapyMOA}: Efficient machine learning for data streams in python,
  2025.

\bibitem{RandomSubSpaces}
Heitor Gomes, Jesse Read, and Albert Bifet.
\newblock Streaming random patches for evolving data stream classification.
\newblock In {\em ICDM'19: IEEE International Conference on Data Mining}, pages
  240--249. IEEE, 2019.

\bibitem{gomes2019machine}
Heitor Gomes, Jesse Read, Albert Bifet, Jean~Paul Barddal, and Joao Gama.
\newblock Machine learning for streaming data: state of the art, challenges,
  and opportunities.
\newblock {\em ACM SIGKDD Explorations Newsletter}, 21(2):6--22, 2019.

\bibitem{DeepLearningBook}
Ian Goodfellow, Yoshua Bengio, and Aaron Courville.
\newblock {\em Deep Learning}.
\newblock The MIT Press, 2016.

\bibitem{grzenda2020delayed}
Maciej Grzenda, Heitor Gomes, and Albert Bifet.
\newblock Delayed labelling evaluation for data streams.
\newblock {\em Data Mining and Knowledge Discovery}, 34(5):1237--1266, 2020.

\bibitem{guzella2009review}
Thiago Guzella and Walmir Caminhas.
\newblock A review of machine learning approaches to spam filtering.
\newblock {\em Expert Systems with Applications}, 36(7):10206--10222, 2009.

\bibitem{Electricity0}
Michael Harries.
\newblock Splice-2 comparative evaluation: Electricity pricing.
\newblock Technical Report, 1999.
\newblock University of New South Wales.

\bibitem{Harries98}
Michael Harries, Claude Sammut, and K.im Horn.
\newblock Extracting hidden context.
\newblock {\em Machine Learning}, 32(2):101--126, 1998.

\bibitem{Haug22}
Johannes Haug, Alexander Braun, Stefan Z\"{u}rn, and Gjergji Kasneci.
\newblock Change detection for local explainability in evolving data streams.
\newblock In {\em Proceedings of the 31st ACM International Conference on
  Information \& Knowledge Management}, CIKM '22, page 706–716, 2022.

\bibitem{haug2022dynamic}
Johannes Haug, Klaus Broelemann, and Gjergji Kasneci.
\newblock Dynamic model tree for interpretable data stream learning.
\newblock In {\em 2022 IEEE 38th International Conference on Data Engineering
  (ICDE)}, pages 2562--2574. IEEE, 2022.

\bibitem{haug2021learning}
Johannes Haug and Gjergji Kasneci.
\newblock Learning parameter distributions to detect concept drift in data
  streams.
\newblock In {\em 2020 25th International Conference on Pattern Recognition
  (ICPR)}, pages 9452--9459. IEEE, 2021.

\bibitem{HolmesKP05}
Geoffrey Holmes, Richard Kirkby, and Bernhard Pfahringer.
\newblock {Stress-testing Hoeffding trees}.
\newblock In {\em 9th European Conference on Principles and Practice of
  Knowledge Discovery in Databases (PKDD '05)}, pages 495--502, 2005.

\bibitem{hou2017one}
Chenping Hou and Zhi-Hua Zhou.
\newblock One-pass learning with incremental and decremental features.
\newblock {\em IEEE transactions on pattern analysis and machine intelligence},
  40(11):2776--2792, 2017.

\bibitem{Hyndman_book}
R.J. Hyndman and G.~Athanasopoulos.
\newblock {\em Forecasting: principles and practice}.
\newblock OTexts: Melbourne, Australia, 2 edition, 2018.

\bibitem{ibarz2021train}
Julian Ibarz, Jie Tan, Chelsea Finn, Mrinal Kalakrishnan, Peter Pastor, and
  Sergey Levine.
\newblock How to train your robot with deep reinforcement learning: lessons we
  have learned.
\newblock {\em The International Journal of Robotics Research},
  40(4-5):698--721, 2021.

\bibitem{iqbal2024anomaly}
Amjad Iqbal, Rashid Amin, Faisal~S Alsubaei, and Abdulrahman Alzahrani.
\newblock Anomaly detection in multivariate time series data using deep
  ensemble models.
\newblock {\em Plos one}, 19(6):e0303890, 2024.

\bibitem{jordan1998notes}
Michael Jordan.
\newblock Notes on recursive least squares.
\newblock Technical Report, University of California, Berkeley, 1998.
\newblock \url{www.cs.berkeley.edu/~jordan/courses/294-fall98/readings/rls.ps}.

\bibitem{kalman1960new}
Rudolph~Emil Kalman.
\newblock A new approach to linear filtering and prediction problems.
\newblock 1960.

\bibitem{kanoun2016big}
Karim Kanoun, Cem Tekin, David Atienza, and Mihaela Van Der~Schaar.
\newblock Big-data streaming applications scheduling based on staged
  multi-armed bandits.
\newblock {\em IEEE Transactions on Computers}, 65(12):3591--3605, 2016.

\bibitem{khan2022nowcasting}
Mohammad Arafat~Ali Khan, Chandra Bhushan, Vadlamani Ravi, Vangala~Sarveswara
  Rao, and Shiva~Shankar Orsu.
\newblock Nowcasting the financial time series with streaming data analytics
  under apache spark.
\newblock {\em arXiv preprint arXiv:2202.11820}, 2022.

\bibitem{kirkpatrick2017overcoming}
James Kirkpatrick, Razvan Pascanu, Neil Rabinowitz, Joel Veness, Guillaume
  Desjardins, Andrei~A Rusu, Kieran Milan, John Quan, Tiago Ramalho, Agnieszka
  Grabska-Barwinska, et~al.
\newblock Overcoming catastrophic forgetting in neural networks.
\newblock {\em Proceedings of the national academy of sciences},
  114(13):3521--3526, 2017.

\bibitem{kolter2005using}
Zico Kolter and Marcus Maloof.
\newblock Using additive expert ensembles to cope with concept drift.
\newblock In {\em Proc. of the 22nd int. conf. on Machine learning}, pages
  449--456, 2005.

\bibitem{ksieniewicz2019data}
Pawe{\l} Ksieniewicz, Micha{\l} Wo{\'z}niak, Bogus{\l}aw Cyganek, Andrzej
  Kasprzak, and Krzysztof Walkowiak.
\newblock Data stream classification using active learned neural networks.
\newblock {\em Neurocomputing}, 353:74--82, 2019.

\bibitem{kudithipudi2022biological}
Dhireesha Kudithipudi, Mario Aguilar-Simon, Jonathan Babb, Maxim Bazhenov,
  Douglas Blackiston, Josh Bongard, Andrew~P Brna, Suraj Chakravarthi~Raja,
  Nick Cheney, Jeff Clune, et~al.
\newblock Biological underpinnings for lifelong learning machines.
\newblock {\em Nature Machine Intelligence}, 4(3):196--210, 2022.

\bibitem{Kuncheva_book}
Ludmila Kuncheva.
\newblock {\em Combining Pattern Classifiers. Methods and Algorithms}.
\newblock 2 edition, 2014.

\bibitem{Kuncheva09}
Ludmila Kuncheva and Indre Zliobaite.
\newblock On the window size for classification in changing environments.
\newblock {\em Intelligent Data Analysis}, 13(6):861--872, 2009.

\bibitem{lim2021time}
Bryan Lim and Stefan Zohren.
\newblock Time-series forecasting with deep learning: a survey.
\newblock {\em Philosophical Transactions of the Royal Society A},
  379(2194):20200209, 2021.

\bibitem{lobo2020spiking}
Jorge Lobo, Javier Del~Ser, Albert Bifet, and Nikola Kasabov.
\newblock Spiking neural networks and online learning: An overview and
  perspectives.
\newblock {\em Neural Networks}, 121:88--100, 2020.

\bibitem{Lu19}
J.~Lu, A.~Liu, F.~Dong, F.~Gu, J.~Gama, and G.~Zhang.
\newblock Learning under concept drift: A review.
\newblock {\em IEEE Transactions on Knowledge and Data Engineering},
  31(12):2346--2363, 2019.

\bibitem{lu2018learning}
Jie Lu, Anjin Liu, Fan Dong, Feng Gu, Joao Gama, and Guangquan Zhang.
\newblock Learning under concept drift: A review.
\newblock {\em IEEE Trans. on Knowledge and Data Engineering},
  31(12):2346--2363, 2018.

\bibitem{lughofer2017line}
Edwin Lughofer.
\newblock On-line active learning: A new paradigm to improve practical
  useability of data stream modeling methods.
\newblock {\em Information Sciences}, 415:356--376, 2017.

\bibitem{luo2020appraisal}
Yong Luo, Liancheng Yin, Wenchao Bai, and Keming Mao.
\newblock An appraisal of incremental learning methods.
\newblock {\em Entropy}, 22(11):1190, 2020.

\bibitem{maggini2019learning}
Marco Maggini, Giuseppe Marra, Stefano Melacci, and Andrea Zugarini.
\newblock Learning in text streams: Discovery and disambiguation of entity and
  relation instances.
\newblock {\em IEEE Transactions on Neural Networks and Learning Systems},
  31(11):4475--4486, 2019.

\bibitem{ExtremelyFastDecisionTrees}
Chaitanya Manapragada, Geoffrey Webb, and Mahsa Salehi.
\newblock Extremely fast decision tree.
\newblock In {\em Proc. of the 24th ACM SIGKDD Int. Conf. on Knowledge
  Discovery \& Data Mining}, page 1953–1962, 2018.

\bibitem{marfoq2023federated}
Othmane Marfoq, Giovanni Neglia, Laetitia Kameni, and Richard Vidal.
\newblock Federated learning for data streams.
\newblock In {\em International Conference on Artificial Intelligence and
  Statistics}, pages 8889--8924. PMLR, 2023.

\bibitem{matuszyk2013correcting}
Pawel Matuszyk, Georg Krempl, and Myra Spiliopoulou.
\newblock Correcting the usage of the {H}oeffding inequality in stream mining.
\newblock In {\em Int. Symp. on Intelligent Data Analysis}, pages 298--309,
  2013.

\bibitem{mawuli2023semi}
Cobbinah Mawuli, Jay Kumar, Ebenezer Nanor, Shangxuan Fu, Liangxu Pan, Qinli
  Yang, Wei Zhang, and Junming Shao.
\newblock Semi-supervised federated learning on evolving data streams.
\newblock {\em Information Sciences}, 643:119235, 2023.

\bibitem{melgar2021nearest}
Laura Melgar-Garc{\'\i}a, David Guti{\'e}rrez-Avil{\'e}s, Cristina
  Rubio-Escudero, and A~Troncoso.
\newblock Nearest neighbors-based forecasting for electricity demand time
  series in streaming.
\newblock In {\em Advances in Artificial Intelligence: 19th Conference of the
  Spanish Association for Artificial Intelligence, CAEPIA 2020/2021,
  M{\'a}laga, Spain, September 22--24, 2021, Proceedings 19}, pages 185--195.
  Springer, 2021.

\bibitem{moerland2023model}
Thomas Moerland, Joost Broekens, Aske Plaat, Catholijn Jonker, et~al.
\newblock Model-based reinforcement learning: A survey.
\newblock {\em Foundations and Trends{\textregistered} in Machine Learning},
  16(1):1--118, 2023.

\bibitem{FastAndSlow}
Jacob Montiel, Albert Bifet, Viktor Losing, Jesse Read, and Talel Abdessalem.
\newblock Learning fast and slow - {A} unified batch/stream framework.
\newblock In {\em {IEEE BigData} Int. Conf. on Big Data}, pages 1065--1072,
  2018.

\bibitem{moran2016epidemic}
Kelly Moran, Geoffrey Fairchild, Nicholas Generous, Kyle Hickmann, Dave Osthus,
  Reid Priedhorsky, James Hyman, and Sara Del~Valle.
\newblock Epidemic forecasting is messier than weather forecasting: The role of
  human behavior and internet data streams in epidemic forecast.
\newblock {\em The Journal of infectious diseases}, 214(suppl\_4):S404--S408,
  2016.

\bibitem{moreira2018classifying}
Denis Moreira~dos Reis, Andr{\'e} Maletzke, Diego~F Silva, and Gustavo~EAPA
  Batista.
\newblock Classifying and counting with recurrent contexts.
\newblock In {\em Proceedings of the 24th ACM SIGKDD International Conference
  on Knowledge Discovery \& Data Mining}, pages 1983--1992, 2018.

\bibitem{CUMSUM}
S.~Muthukrishnan, Eric van~den Berg, and Yihua Wu.
\newblock Sequential change detection on data streams.
\newblock In {\em Workshops Proceedings of the 7th {IEEE} International
  Conference on Data Mining}, pages 551--550, 2007.

\bibitem{nouvellet2021reduction}
Pierre Nouvellet, Sangeeta Bhatia, Anne Cori, Kylie~EC Ainslie, Marc Baguelin,
  Samir Bhatt, Adhiratha Boonyasiri, Nicholas~F Brazeau, Lorenzo Cattarino,
  Laura~V Cooper, et~al.
\newblock Reduction in mobility and covid-19 transmission.
\newblock {\em Nature communications}, 12(1):1090, 2021.

\bibitem{openai_gpt_2024}
{OpenAI}.
\newblock {GPT} models and their capabilities, 2024.
\newblock Accessed: 2025-04-24.

\bibitem{Pan09}
S.~Pan and Q.~Yang.
\newblock A survey on transfer learning.
\newblock {\em IEEE Transactions on Knowledge and Data Engineering},
  22(10):1345--1359, 2010.

\bibitem{parisi2019continual}
German Parisi, Ronald Kemker, Jose Part, Christopher Kanan, and Stefan Wermter.
\newblock Continual lifelong learning with neural networks: A review.
\newblock {\em Neural networks}, 113:54--71, 2019.

\bibitem{pinckaers2020streaming}
Hans Pinckaers, Bram Van~Ginneken, and Geert Litjens.
\newblock Streaming convolutional neural networks for end-to-end learning with
  multi-megapixel images.
\newblock {\em IEEE transactions on pattern analysis and machine intelligence},
  44(3):1581--1590, 2020.

\bibitem{Plasse16}
Joshua Plasse and Neil Adams.
\newblock Handling delayed labels in temporally evolving data streams.
\newblock In {\em IEEE Int. Conf. on Big Data}, pages 2416--2424, 2016.

\bibitem{AlbanTurbines}
Alban Puech and Jesse Read.
\newblock An improved yaw control algorithm for wind turbines via reinforcement
  learning.
\newblock In {\em ECML-PKDD 2022: 33rd European Conference on Machine
  Learning}, pages 614--630. Springer Nature Switzerland, 2023.
\newblock ADS Track.

\bibitem{quinonero2022dataset}
Joaquin Qui{\~n}onero-Candela, Masashi Sugiyama, Anton Schwaighofer, and Neil~D
  Lawrence.
\newblock {\em Dataset shift in machine learning}.
\newblock Mit Press, 2022.

\bibitem{rauba2024self}
Paulius Rauba, Nabeel Seedat, Krzysztof Kacprzyk, and Mihaela van~der Schaar.
\newblock Self-healing machine learning: A framework for autonomous adaptation
  in real-world environments.
\newblock {\em Advances in Neural Information Processing Systems},
  37:42225--42267, 2024.

\bibitem{Provocative1}
Jesse Read.
\newblock Concept-drifting data streams are time series; the case for
  continuous adaptation.
\newblock {\em arXiv preprint arXiv:1810.02266}, 2018.

\bibitem{IDA2012}
Jesse Read, Albert Bifet, Bernhard Pfahringer, and Geoff Holmes.
\newblock Batch-incremental versus instance-incremental learning in dynamic and
  evolving data.
\newblock In {\em IDA: 11th Int. Symp. on Advances in Intelligent Data
  Analysis}, pages 313--323, 2012.

\bibitem{ChainInTime}
Jesse Read, Luca Martino, and Jaakko Hollmen.
\newblock Multi-label methods for prediction with sequential data.
\newblock {\em Pattern Recognition}, 63(March):45--55, 2017.

\bibitem{Provocative2}
Jesse Read, Ricardo Rios, Tatiane Nogueira, and Rodrigo Mello.
\newblock Data streams are time series: Challenging assumptions.
\newblock In {\em 9th Brazilian Conference on Intelligent Systems (BRACIS)},
  pages 529--543, 2020.

\bibitem{ESR}
Jesse Read, Nikolaos Tziortziotis, and Michalis Vazirgiannis.
\newblock Error-space representations for multi-dimensional data-streams.
\newblock {\em Pattern Analysis and Applications}, 22(3):1211--1220, 2019.

\bibitem{pmlr-v94-roseberry18a}
Martha Roseberry and Alberto Cano.
\newblock Multi-label {kNN} classifier with self adjusting memory for drifting
  data streams.
\newblock In {\em Proc. of the 2nd Int. Workshop on Learning with Imbalanced
  Domains: Theory and Applications}, volume~94, pages 23--37, 2018.

\bibitem{roseberry2021self}
Martha Roseberry, Bartosz Krawczyk, Youcef Djenouri, and Alberto Cano.
\newblock Self-adjusting k nearest neighbors for continual learning from
  multi-label drifting data streams.
\newblock {\em Neurocomputing}, 442:10--25, 2021.

\bibitem{rutkowski2013decision}
Leszek Rutkowski, Maciej Jaworski, Lena Pietruczuk, and Piotr Duda.
\newblock Decision trees for mining data streams based on the gaussian
  approximation.
\newblock {\em IEEE Transactions on Knowledge and Data Engineering},
  26(1):108--119, 2013.

\bibitem{sanchez2019data}
Javier~J S{\'a}nchez-Medina, Juan~Antonio Guerra-Montenegro, David
  S{\'a}nchez-Rodr{\'\i}guez, Itziar~G Alonso-Gonz{\'a}lez, and Juan~L
  Navarro-Mesa.
\newblock Data stream mining applied to maximum wind forecasting in the canary
  islands.
\newblock {\em Sensors}, 19(10):2388, 2019.

\bibitem{sculley2007online}
D~Sculley.
\newblock Online active learning methods for fast label-efficient spam
  filtering.
\newblock In {\em CEAS}, volume~7, page 143, 2007.

\bibitem{sebastiao2009study}
Raquel Sebastiao and Joao Gama.
\newblock A study on change detection methods.
\newblock In {\em Progress in artificial intelligence, 14th Portuguese
  conference on artificial intelligence, EPIA}, pages 12--15, 2009.

\bibitem{Shimodaira00}
Hidetoshi Shimodaira.
\newblock Improving predictive inference under covariate shift by weighting the
  log-likelihood function.
\newblock {\em Journal of Statistical Planning and Inference}, 90(2):227--244,
  2000.

\bibitem{silva2024streaming}
Tiago Silva, Daniel~Augusto de~Souza, and Diego Mesquita.
\newblock Streaming bayes gflownets.
\newblock {\em Advances in Neural Information Processing Systems},
  37:27153--27177, 2024.

\bibitem{singharoy2018tree}
Kunal SinghaRoy.
\newblock {\em Tree-Augmented Na{\"\i}ve Bayes Methods for Real-Time Training
  and Classification of Streaming Data}.
\newblock PhD thesis, The Florida State University, 2018.

\bibitem{soemers2018adapting}
Dennis Soemers, Tim Brys, Kurt Driessens, Mark Winands, and Ann Now{\'e}.
\newblock Adapting to concept drift in credit card transaction data streams
  using contextual bandits and decision trees.
\newblock In {\em Proceedings of the AAAI conference on artificial
  intelligence}, volume~32, 2018.

\bibitem{suganthi2012energy}
L~Suganthi and Anand Samuel.
\newblock Energy models for demand forecasting—a review.
\newblock {\em Renewable and sustainable energy reviews}, 16(2):1223--1240,
  2012.

\bibitem{sun2018concept}
Yu~Sun, Ke~Tang, Zexuan Zhu, and Xin Yao.
\newblock Concept drift adaptation by exploiting historical knowledge.
\newblock {\em IEEE transactions on neural networks and learning systems},
  29(10):4822--4832, 2018.

\bibitem{RLBook2018}
Richard~S. Sutton and Andrew~G. Barto.
\newblock {\em Reinforcement Learning: An Introduction}.
\newblock The MIT Press, second edition, 2018.

\bibitem{AdversarialOnline}
Yufei Tao and Shangqi Lu.
\newblock From online to non-i.i.d. batch learning.
\newblock In {\em Proceedings of the 26th ACM SIGKDD International Conference
  on Knowledge Discovery and Data Mining}, KDD '20, page 328–337, 2020.

\bibitem{taylor2009transfer}
Matthew~E Taylor and Peter Stone.
\newblock Transfer learning for reinforcement learning domains: A survey.
\newblock {\em Journal of Machine Learning Research}, 10(7), 2009.

\bibitem{THEODORIDIS20159}
Sergios Theodoridis.
\newblock Chapter 2 - probability and stochastic processes.
\newblock In Sergios Theodoridis, editor, {\em Machine Learning}, pages 9--51.
  Academic Press, Oxford, 2015.

\bibitem{Tsymbal04}
Alexey Tsymbal.
\newblock The problem of concept drift definitions and related work.
\newblock Tech. Rep. Department of Computer Science, Trinity College, Dublin.,
  2004.

\bibitem{van2022three}
Gido van~de Ven, Tinne Tuytelaars, and Andreas Tolias.
\newblock Three types of incremental learning.
\newblock {\em Nature Machine Intelligence}, pages 1--13, 2022.

\bibitem{vela2022temporal}
Daniel Vela, Andrew Sharp, Richard Zhang, Trang Nguyen, An~Hoang, and Oleg~S
  Pianykh.
\newblock Temporal quality degradation in ai models.
\newblock {\em Scientific Reports}, 12(1):11654, 2022.

\bibitem{wald2021calibration}
Yoav Wald, Amir Feder, Daniel Greenfeld, and Uri Shalit.
\newblock On calibration and out-of-domain generalization.
\newblock {\em Advances in neural information processing systems},
  34:2215--2227, 2021.

\bibitem{webb2016characterizing}
Geoffrey Webb, Roy Hyde, Hong Cao, Hai~Long Nguyen, and François Petitjean.
\newblock Characterizing concept drift.
\newblock {\em Data Mining and Knowledge Discovery}, 30(4):964--994, 2016.

\bibitem{Widmer96}
Gerhard Widmer and Miroslav Kubat.
\newblock Learning in the presence of concept drift and hidden contexts.
\newblock {\em Mach. Learn.}, 23(1):69–101, 1996.

\bibitem{Widrow59}
Bernard Widrow and Marcian~E Hoff.
\newblock Adaptive switching circuits.
\newblock In {\em IRE WESCON CONVENTION RECORD}, 1960.

\bibitem{wolf2017we}
Marty Wolf, Keith Miller, and Frances Grodzinsky.
\newblock Why we should have seen that coming: comments on microsoft's tay
  "experiment," and wider implications.
\newblock {\em {ACM} {SIGCAS} Computers and Society}, 47(3):54--64, 2017.

\bibitem{Wolpert97}
David Wolpert and William Macready.
\newblock No free lunch theorems for optimization.
\newblock {\em IEEE Transactions on Evolutionary Computation}, 1(1):67--82,
  1997.

\bibitem{xing2012early}
Zhenhui Xing, Jian Pei, and Philip~S. Yu.
\newblock Early classification on time series.
\newblock {\em Knowledge and information systems}, 31(1):105--127, 2012.

\bibitem{yao2024survey}
Yifan Yao, Jinhao Duan, Kaidi Xu, Yuanfang Cai, Zhibo Sun, and Yue Zhang.
\newblock A survey on large language model (llm) security and privacy: The
  good, the bad, and the ugly.
\newblock {\em High-Confidence Computing}, page 100211, 2024.

\bibitem{you2021learning}
Xiaoyang You, Mengdi Zhang, Daxin Ding, Fuli Feng, and Yongfeng Huang.
\newblock Learning to learn the future: Modeling concept drifts in time series
  prediction.
\newblock In {\em Proc. of the 30th ACM Int. Conf. on Information \& Knowledge
  Management}, pages 2434--2443, 2021.

\bibitem{zare2024survey}
Maryam Zare, Parham~M Kebria, Abbas Khosravi, and Saeid Nahavandi.
\newblock A survey of imitation learning: Algorithms, recent developments, and
  challenges.
\newblock {\em IEEE Transactions on Cybernetics}, 2024.

\bibitem{zhai2017classification}
Tingting Zhai, Yang Gao, Hao Wang, and Longbing Cao.
\newblock Classification of high-dimensional evolving data streams via a
  resource-efficient online ensemble.
\newblock {\em Data Mining and Knowledge Discovery}, 31:1242--1265, 2017.

\bibitem{zhang2017deeper}
Shangtong Zhang and Richard~S Sutton.
\newblock A deeper look at experience replay.
\newblock {\em arXiv preprint arXiv:1712.01275}, 2017.

\bibitem{ziffer2021kalman}
Giacomo Ziffer, Alessio Bernardo, Emanuele Della~Valle, and Albert Bifet.
\newblock Kalman filtering for learning with evolving data streams.
\newblock In {\em 2021 IEEE International Conference on Big Data (Big Data)},
  pages 5337--5346. IEEE, 2021.

\bibitem{Zliobaite_thesis}
Indre Zliobaite.
\newblock {\em Adaptive training set formation}.
\newblock PhD thesis, Vilnius University, 2010.

\bibitem{Zliobaite10detectable}
Indre Zliobaite.
\newblock Change with delayed labeling: When is it detectable?
\newblock In {\em IEEE International Conference on Data Mining Workshops},
  pages 843--850, 2010.

\bibitem{Zliobaite11}
Indre Zliobaite.
\newblock Combining similarity in time and space for training set formation
  under concept drift.
\newblock {\em Intelligent Data Analysis}, 15(4):589--611, 2011.

\bibitem{Zliobaite13electricity}
Indre Zliobaite.
\newblock How good is the electricity benchmark for evaluating concept drift
  adaptation.
\newblock {\em arXiv:1301.3524}, 2013.

\bibitem{Active}
Indre Zliobaite, Albert Bifet, Geoffrey Holmes, and Bernard Pfahringer.
\newblock {MOA} concept drift active learning strategies for streaming data.
\newblock {\em Journal of Machine Learning Research - Proceedings Track},
  17:48--55, 2011.

\bibitem{Our2015Paper}
Indre Zliobaite, Albert Bifet, Jesse Read, Bernard Pfahringer, and Geoffrey
  Holmes.
\newblock Evaluation methods and decision theory for classification of
  streaming data with temporal dependence.
\newblock {\em Machine Learning}, 98(3):455--482, 2014.

\bibitem{vzliobaite2015towards}
Indre Zliobaite, Marcin Budka, and Frederic Stahl.
\newblock Towards cost-sensitive adaptation: When is it worth updating your
  predictive model?
\newblock {\em Neurocomputing}, 150:240--249, 2015.

\bibitem{HistoricalPerspective}
Indre Zliobaite and Jesse Read.
\newblock Data streams: A historical perspective.
\newblock {\em arXiv preprint}, 2023.

\end{thebibliography}

\end{document}